 \documentclass[twocolumn,twoside]{IEEEtran}
\usepackage{amsmath,amssymb,amsthm,wasysym,epsfig,color,subfigure,graphicx,epstopdf,url,bm,nicefrac,tcolorbox}
\usepackage{caption}
\usepackage{amsmath,graphicx,bm,accents}

\usepackage{multicol}
\usepackage{multirow}

\usepackage{algorithm}
\usepackage{algorithmic}
\allowdisplaybreaks

\usepackage{hyperref}

\hypersetup{colorlinks=false}

\usepackage{accents}
\makeatletter
\def\wid{\check{{\cc@style\underline{\mskip9.5mu}}}}
\def\Wideubar{\underaccent{{\cc@style\underline{\mskip6mu}}}}
\makeatother

\makeatletter
\def\wideubar{\underaccent{{\cc@style\underline{\mskip9.5mu}}}}
\def\Wideubar{\underaccent{{\cc@style\underline{\mskip6mu}}}}
\makeatother

\makeatletter
\def\widebar{\accentset{{\cc@style\underline{\mskip9.5mu}}}}
\def\Widebar{\accentset{{\cc@style\underline{\mskip6mu}}}}
\makeatother

\newtheorem{theorem}{Theorem}

\newtheorem{assumption}{Assumption}
\theoremstyle{remark}\newtheorem{remark}{Remark}

\begin{document}
\title{Nonlinear Dimensionality Reduction for Discriminative Analytics of Multiple Datasets
}

%\markboth{IEEE TRANSACTIONS ON SIGNAL  PROCESSING (Accepted, \today)}{}

%\author{
%
%	\thanks{
%	Manuscript received December 11, 2016; revised April 5, 2017 and August 25, 2017; accepted October  29, 2017. Date of publication DATE; date of current version DATE. 
%	
%
%}
%}

\author{Jia Chen,
	Gang Wang,~\IEEEmembership{Member,~IEEE},
	and 
	Georgios B. Giannakis,~\IEEEmembership{Fellow,~IEEE}
	\thanks{
%		Manuscript received May 13, 2018; revised August 24, 2018 and October 17, 2018; accepted November 22, 2018. Date of publication xxx, 2018; date of current version xxx, 2018.  The associate editor coordinating the review of this manuscript and approving it for publication was Prof. Byonghyo Shim. 
		This work was supported in part by NSF grants 1711471, 1514056, and the NIH grant 1R01GM104975-01. 
		This paper was presented in part at the 43rd IEEE Intl. Conf. on Acoustics, Speech, and Signal Processing, Calgary, Canada, April 15-20, 2018 \cite{2018dpca}, and the 52nd Asilomar Conf. on Signals, Systems, and Computers, Pacific Grove, CA, October 28-31, 2018 \cite{asilomar2018cwg}. 

	The authors are with the Digital Technology Center and the Department of Electrical and Computer Engineering, University of Minnesota, Minneapolis, MN 55455 USA (e-mail: chen5625@umn.edu, gangwang@umn.edu, georgios@umn.edu). }
}

\maketitle

\allowdisplaybreaks

\begin{abstract}
Principal component analysis (PCA) is widely used for feature extraction and dimensionality reduction, with documented merits in diverse tasks involving high-dimensional data. PCA copes with one dataset at a time, but it is challenged when 
it comes to 
analyzing multiple datasets jointly. In certain data science settings however, one is often interested in
 extracting the most discriminative information from one dataset of particular interest (a.k.a. target data) relative to the other(s) (a.k.a. background data).
 To this end, this paper puts forth a novel approach, termed \emph{discriminative} (d) PCA, for such discriminative analytics of multiple datasets. Under certain conditions, dPCA is proved to be 
least-squares
  optimal in recovering the latent subspace vector unique to the target data relative to background data. 
To account for nonlinear data correlations, (linear) dPCA models for one or multiple background datasets are generalized through kernel-based learning. Interestingly, all dPCA variants admit an analytical solution obtainable with a single (generalized) eigenvalue decomposition.
Finally, substantial dimensionality reduction tests using synthetic and real datasets are provided to corroborate the merits of the proposed methods.

\end{abstract}


\section{Introduction}\label{sec:intro}
Principal component analysis (PCA) is  
the ``workhorse'' method
  for dimensionality reduction and feature extraction. It finds well-documented applications, including  statistics, bioinformatics, genomics, quantitative finance, and engineering, to name just a few. The goal of PCA is to obtain low-dimensional representations for high-dimensional data, while preserving most of the high-dimensional data variance \cite{1901pca}.

Yet, various practical scenarios involve \emph{multiple} datasets, in which one is tasked with extracting the most discriminative information of one target dataset relative to others. 
For instance, consider two gene-expression measurement datasets of volunteers from across different geographical  areas and genders: the first dataset collects gene-expression levels of cancer patients, considered here as the 
 \emph{target data}, 
while the second contains levels  from healthy individuals corresponding here to our 
 \emph{background data}. The  goal is to identify molecular subtypes of cancer within cancer patients.
Performing PCA on either the target data or the target together with background data is likely to yield principal components (PCs) that correspond to
the background information common to both datasets (e.g., the demographic patterns and genders) \cite{1998background}, rather than the PCs uniquely describing the subtypes of cancer.
Albeit simple to comprehend and practically relevant, such discriminative data analytics has not been thoroughly addressed.

Generalizations of PCA include kernel (K) PCA \cite{kpca,2017kpca}, graph PCA \cite{proc2018gsk}, 
$\ell_1$-PCA \cite{2018l1pca}, 
 robust PCA \cite{jstsp2016shahid},
multi-dimensional scaling \cite{mds}, 
locally linear embedding \cite{lle}, 
Isomap \cite{2000isomap}, and Laplacian eigenmaps \cite{2003eigenmap}. Linear discriminant analysis (LDA) is a
\emph{supervised} classifier of linearly projected reduced dimensionality data vectors. It is designed so that linearly projected training vectors (meaning \emph{labeled} data) of the same class stay as close as possible, while projected data of different classes are positioned as far as possible \cite{1933lda}. 
Other discriminative methods include 
	re-constructive and discriminative subspaces \cite{fidler2006combining},
	 discriminative vanishing component analysis \cite{hou2016discriminative}, 
	 and kernel LDA \cite{mika1999fisher},
	  which similar to LDA rely on labeled data.
Supervised PCA looks for orthogonal projection vectors so that the dependence of projected vectors from 
	one dataset on the other dataset is maximized \cite{barshan2011supervised}.

Multiple-factor analysis, an extension of PCA to deal with multiple datasets, is implemented in two steps: S1) normalize each dataset by the largest eigenvalue of 
	its sample covariance matrix; and, S2) perform PCA on the combined dataset of all normalized ones \cite{abdi2013multiple}.
On the other hand, canonical correlation analysis is widely employed for analyzing multiple datasets \cite{1936cca,2018cwsggcca,2018gmcca}, but its goal is to extract the shared low-dimensional structure. 
The recent proposal called contrastive (c) PCA aims at extracting contrastive information between two datasets \cite{2017cpca}, by searching for directions along which the target data variance is large while that of 
 the background data  is small. Carried out using the singular value decomposition (SVD), cPCA can  reveal dataset-specific information often missed by standard PCA if the involved hyper-parameter is properly selected.  
Though possible 
 to automatically choose
the best hyper-parameter from a list of candidate values, performing SVD multiple
times can be computationally cumbersome in large-scale feature extraction settings.

Building on but going beyond cPCA, this paper starts by developing a novel approach, termed discriminative (d) PCA, for discriminative analytics of \emph{two} datasets. dPCA looks for linear projections (as in LDA) but of \emph{unlabeled}
 data vectors, by \textcolor{black}{maximizing the variance of projected  target data while minimizing that of background data. This  leads to a \emph{ratio trace} maximization formulation,}
 and also justifies our chosen term \emph{discriminative  PCA}. Under certain conditions, dPCA is proved to be least-squares (LS) optimal in the sense that it reveals PCs specific to the target data relative to background data. Different from 
 cPCA, dPCA is parameter free, and it requires a single generalized eigendecomposition, lending itself favorably to large-scale discriminative data analytics. 
  However, real data vectors 
   often exhibit nonlinear correlations, rendering dPCA inadequate for complex practical setups. To this end, nonlinear dPCA is developed via kernel-based learning. Similarly, the solution of KdPCA can be provided analytically in terms of generalized eigenvalue decompositions. As the complexity of KdPCA grows only linearly with the dimensionality of data vectors, KdPCA is preferable over dPCA for discriminative analytics of high-dimensional data.

dPCA is further extended to handle multiple (more than two) background datasets. Multi-background (M) dPCA is developed to extract low-dimensional discriminative structure unique to the target data but not to \emph{multiple} sets of background data. This becomes possible by maximizing  \textcolor{black}{the variance of projected 
	 target data while minimizing the sum of variances of all projected  
	  background data.}
At last, kernel (K) MdPCA is put forth to account for nonlinear data correlations.

%The rest of this paper is organized as follows. Upon reviewing the prior art in Sec. \ref{sec:preli}, linear dPCA is dealt with in Sec \ref{sec:dpca}. Optimality of dPCA is established in Sec. \ref{sec:optim}. Accounting for nonlinearities, KdPCA is advocated in Sec. \ref{sec:kdpca}. Generalizing their single-background variants, multi-background (M) dPCA and KMdPCA models are discussed in Sec. \ref{sec:mdpca}. Numerical tests are reported in Sec. \ref{sec:simul}, while the paper is concluded with  the  research outlook in Sec. \ref{sec:concl}.

\emph{Notation}: Bold uppercase (lowercase) letters denote matrices (column vectors).
Operators $(\cdot)^{\top}$,
$(\cdot)^{-1}$, and ${\rm Tr}(\cdot)$ denote matrix transposition, 
inverse, and trace, respectively; 
$\|\mathbf{a}\|_2 $ is the $\ell_2$-norm of vector $\mathbf{a}$;
%  $\mathbf{A}\succ \mathbf{0}$ means that symmetric matrix $\mathbf{A}$ is positive definite; 
  ${\rm diag}(\{a_i\}_{i=1}^m)$ is a diagonal matrix holding elements $\{a_i\}_{i=1}^m$ on its main diagonal; 
  $\mathbf{0}$ denotes all-zero vectors or matrices;
  and $\mathbf{I}$ represents identity matrices of suitable dimensions.

\section{Preliminaries and Prior Art}\label{sec:preli}
Consider two datasets, namely a target dataset $\{\mathbf{x}_i\in\mathbb{R}^D\}_{i=1}^m$ that we are interested in analyzing, and a background dataset $\{\mathbf{y}_j\in\mathbb{R}^D\}_{j=1}^n$ that contains latent background-related vectors also present  in the target data. Generalization to multiple background datasets will be presented in Sec. \ref{sec:mdpca}. 
Assume without loss of generality  that both datasets are centered; in other words, 
\textcolor{black}{the sample mean $m^{-1}\sum_{i=1}^m \mathbf{x}_i$ $(n^{-1}\sum_{j=1}^n \mathbf{y}_j)$ has been subtracted from each $\mathbf{x}_i$ $(\mathbf{y}_j)$.}
To motivate our novel approaches 
in subsequent sections, some basics of PCA and cPCA are outlined next.

Standard PCA handles a single dataset at a time. 
It looks for low-dimensional representations $\{\boldsymbol{\chi}_i\in\mathbb{R}^d \}_{i=1}^m$ of $\{\mathbf{x}_i\}_{i=1}^m$ with $d<D$ as linear projections  of $\{\mathbf{x}_i \}_{i=1}^m$ by maximizing the variances of $\{\bm{\chi}_i \}_{i=1}^m$ \cite{1901pca}. Specifically for $d=1$, (linear) PCA yields  ${\chi}_i:=\hat{\mathbf{u}}^\top\mathbf{x}_i$, with the  vector $\hat{\mathbf{u}}\in\mathbb{R}^D$ found by
	\begin{equation}
		\label{eq:pca}
			\hat{\mathbf{u}}:=\arg
			\underset{\mathbf{u}\in\mathbb{R}^D}{\max}
			\quad\mathbf{u}^\top\mathbf{C}_{xx}\mathbf{u}\quad {\rm s.\,to}\quad \mathbf{u}^\top\mathbf{u}=1	
		\end{equation}	
where $\mathbf{C}_{xx}:=(1/m)\sum_{i=1}^m\mathbf{x}_i\mathbf{x}_i^\top\in\mathbb{R}^{D\times D}$ is the sample covariance matrix of $\{\mathbf{x}_i \}_{i=1}^m$. Solving \eqref{eq:pca} yields $\hat{\mathbf{u}}$ as the \textcolor{black}{normalized} eigenvector of $\mathbf{C}_{xx}$ corresponding to the largest eigenvalue.
The resulting projections $\{\chi_i =\hat{\mathbf{u}}^\top\mathbf{x}_i \}_{i=1}^m$ 
\textcolor{black}{constitute}
  the first principal 
  \textcolor{black}{component (PC)} of the target data vectors. 
When $d>1$, PCA looks for  $\{\mathbf{u}_i\in\mathbb{R}^D \}_{i=1}^d$, 
obtained from  the \textcolor{black}{$d$ eigenvectors of $\mathbf{C}_{xx}$ associated with the first $d$ largest eigenvalues sorted in a decreasing order}. As alluded to in Sec. \ref{sec:intro}, PCA applied on $\{\mathbf{x}_i \}_{i=1}^m$ only, or on the combined datasets
 $\{\{\mathbf{x}_i\}_{i=1}^m,\,\{\mathbf{y}_j\}_{j=1}^n \}$ can generally not uncover the discriminative patterns or features of the target data relative to the background data.

On the other hand, the recent cPCA seeks a vector $\mathbf{u}\in\mathbb{R}^D$ along which the target data exhibit large variations while the background
 data exhibit small variations, via solving \cite{2017cpca}
	\begin{subequations}
		\label{eq:cpca}
		\begin{align}
			\underset{\mathbf{u}\in\mathbb{R}^D}{\max}
			\quad&\mathbf{u}^\top\mathbf{C}_{xx}\mathbf{u}-\alpha \mathbf{u}^\top\mathbf{C}_{yy}\mathbf{u}\label{eq:cpcaobj}\\
			{\rm s.\,to}\quad &\mathbf{u}^\top\mathbf{u}=1
		\end{align}		
	\end{subequations}
where $\mathbf{C}_{yy}:=(1/n)\sum_{j=1}^n\mathbf{y}_j\mathbf{y}_j^\top\in\mathbb{R}^{D\times D}$ denotes the sample covariance matrix of $\{\mathbf{y}_j \}_{j=1}^n$, and the hyper-parameter $\alpha\ge 0$ trades off maximizing the target data variance (the first term in \eqref{eq:cpcaobj}) for minimizing the background data variance (the second term). For a given $\alpha$, the solution of \eqref{eq:cpca} is given by the eigenvector of $\mathbf{C}_{xx}-\alpha\mathbf{C}_{yy}$ associated with its largest eigenvalue, along which the obtained data projections constitute  the first contrastive (c) \textcolor{black}{PC}. Nonetheless, there is no rule of thumb for choosing $\alpha$.
\textcolor{black}{A spectral-clustering based algorithm was devised to automatically select $\alpha$ from a list of candidate values \cite{2017cpca}, but its brute-force search is computationally expensive to use in large-scale datasets.}

\section{Discriminative Principal Component Analysis} \label{sec:dpca}
Unlike PCA, LDA is a supervised classification method of linearly projected data at reduced dimensionality.   
It finds those linear projections  that reduce that variation in the same class and increase the separation between classes \cite{1933lda}. This is accomplished by maximizing the ratio of the labeled data variance between classes to that within the classes. 

In a related but unsupervised setup, 
consider we are given a target dataset and a background dataset, and we are tasked with \textcolor{black}{extracting   vectors that are meaningful in representing
	  $\{\mathbf{x}_i\}_{i=1}^m$, but not $\{\mathbf{y}_j\}_{j=1}^n$.}
 A meaningful 
 approach would then be to maximize the ratio of the projected target data variance over that of the background data. Our  \emph{discriminative (d) PCA} approach finds
		\begin{equation}	\label{eq:dpca}
		\textcolor{black}{
			\hat{\mathbf{u}}:=\arg
			\underset{\mathbf{u}\in\mathbb{R}^{D}}{\max}
			\quad\frac{\mathbf{u}^\top\mathbf{C}_{xx}\mathbf{u}}{ \mathbf{u}^\top\mathbf{C}_{yy}\mathbf{u}}}
		\end{equation}
We will term the solution in \eqref{eq:dpca}  discriminant subspace vector,
and the projections $\{\hat{\mathbf{u}}^\top\mathbf{x}_i\}_{i=1}$  the first discriminative (d) \textcolor{black}{PC}. 
Next, we discuss the solution in  \eqref{eq:dpca}.

Using Lagrangian duality theory, the solution in \eqref{eq:dpca} corresponds to the right eigenvector of $\mathbf{C}_{yy}^{-1}\mathbf{C}_{xx}$ associated with the largest eigenvalue. 
To establish this,  note that \eqref{eq:dpca} can be equivalently rewritten as% \textcolor{black}{(see a proof in Appendix \ref{app:equv})}
	\begin{subequations}
\label{eq:dpcafm2}
\begin{align}
	\hat{\mathbf{u}}:=\arg
		\underset{\mathbf{u}\in\mathbb{R}^{D}}{\max}\quad& \mathbf{u}^\top\mathbf{C}_{xx}\mathbf{u}\label{eq:dpcafm2cos}\\
	{\rm s.\,to}\quad& \mathbf{u}^\top\mathbf{C}_{yy}\mathbf{u}=1.\label{eq:dpcafm2con}
\end{align}
	\end{subequations}
	Letting $\lambda$ denote the dual variable associated with the constraint \eqref{eq:dpcafm2con}, the Lagrangian of \eqref{eq:dpcafm2} becomes
\begin{equation}\label{eq:lag}
\mathcal{L}(\mathbf{u};\,\lambda)=\mathbf{u}^\top\mathbf{C}_{xx}\mathbf{u}+\lambda\left(1-\mathbf{u}^\top\mathbf{C}_{yy}\mathbf{u}\right).
\end{equation}
At the optimum $(\hat{\mathbf{u}};\,\hat{\lambda})$, 
the KKT conditions confirm that
\begin{equation}\label{eq:gep}
	\mathbf{C}_{xx}\hat{{\mathbf{u}}}=\hat{\lambda}\mathbf{C}_{yy}\hat{\mathbf{u}}.
\end{equation}
This is a generalized eigen-equation, whose solution $\hat{\mathbf{u}}$ is the generalized eigenvector of $(\mathbf{C}_{xx},\,\mathbf{C}_{yy})$ corresponding to the generalized eigenvalue $\hat{\lambda}$. 
Left-multiplying
  \eqref{eq:gep} by $\hat{\mathbf{u}}^\top$ yields
$	\hat{\mathbf{u}}^\top\mathbf{C}_{xx}\hat{\mathbf{u}}=\hat{\lambda} \hat{\mathbf{u}}^\top\mathbf{C}_{yy}\hat{\mathbf{u}}
$, corroborating that the optimal objective value of \eqref{eq:dpcafm2cos} is attained when $\hat{\lambda}:=\lambda_1$ is the largest generalized eigenvalue. Furthermore, \eqref{eq:gep} can be solved efficiently using well-documented solvers that rely on e.g., Cholesky's factorization \cite{saad1}.

 Supposing further that $\mathbf{C}_{yy}$ is nonsingular
   \eqref{eq:gep}  yields 
  \begin{equation}
  \label{eq:dpcasol}
  \mathbf{C}_{yy}^{-1}\mathbf{C}_{xx}\hat{\mathbf{u}}=\hat{\lambda}\hat{\mathbf{u}}
  \end{equation}
implying that  $\hat{\mathbf{u}}$ in \eqref{eq:dpcafm2} is the right eigenvector of $\mathbf{C}_{yy}^{-1}\mathbf{C}_{xx}$ corresponding to the largest eigenvalue    $\hat{\lambda}=\lambda_1$.

To find multiple ($d\ge 2$) subspace vectors, namely $\{\mathbf{u}_i\in\mathbb{R}^D\}_{i=1}^d$ that form $\mathbf{U}:=[\mathbf{u}_1 \, \cdots \, \mathbf{u}_d]\in\mathbb{R}^{D\times d}$, \textcolor{black}{in \eqref{eq:dpca} with  $\mathbf{C}_{yy}$ being nonsingular,}  can be generalized as follows (cf. \eqref{eq:dpca})
\vspace{4pt}
\textcolor{black}{
	\begin{equation}
		\label{eq:dpcam}	
		\hat{\mathbf{U}}:=\arg	\underset{\mathbf{U}\in\mathbb{R}^{D\times d}}{\max}~
		{\rm Tr}\left[\left(\mathbf{U}^\top\mathbf{C}_{yy}\mathbf{U}\right )^{-1}\mathbf{U}^\top\mathbf{C}_{xx}\mathbf{U}\right].
	\end{equation}
	}

Clearly, \eqref{eq:dpcam} is a \emph{ratio trace} maximization problem; see e.g., \cite{2014mati}, whose  solution
is given in Thm. \ref{the:dpca} (see a proof in \cite[p. 448]{2013fukunaga}).

\begin{theorem}
	\label{the:dpca}
	Given centered data $\{{\mathbf{x}}_i\in\mathbb{R}^{D}\}_{i=1}^m$ and $\{{\mathbf{y}}_j\in\mathbb{R}^{D}\}_{j=1}^n$ with sample covariance matrices $\mathbf{C}_{xx}:=(1/m)\sum_{i=1}^m\mathbf{x}_i\mathbf{x}_i^\top$ and $\mathbf{C}_{yy}:=(1/n)\sum_{j=1}^n\mathbf{y}_j\mathbf{y}_j^\top\succ\mathbf{0}$, the $i$-th column of the dPCA optimal solution $\hat{\mathbf{U}}\in\mathbb{R}^{D\times d}$ in \eqref{eq:dpcam} is given by the right eigenvector of  $\mathbf{C}_{yy}^{-1}\mathbf{C}_{xx}$ associated with the $i$-th largest eigenvalue, where $i=1,\,\ldots,\,d$.
\end{theorem}

Our dPCA for 
discriminative analytics of two datasets is summarized in Alg. \ref{alg:dpca}.
\textcolor{black}{Four} remarks are now in order.

\begin{remark}
Without background data, we have $\mathbf{C}_{yy}=\mathbf{I}$, and dPCA boils down to the standard PCA. 
\end{remark}

\begin{remark}
	Several possible combinations of target and background datasets include:
	 i) measurements from a healthy group $\{\mathbf{y}_j\}$ and a diseased group $\{\mathbf{x}_i\}$, where the former has similar population-level variation 
	 with the latter, but distinct variation 
	 due to subtypes of diseases; ii) before-treatment $\{\mathbf{y}_j\}$ and after-treatment   $\{\mathbf{x}_i\}$ datasets, in which the former contains additive 
	 measurement noise rather than the variation caused by treatment; and iii) signal-free $\{\mathbf{y}_j\}$ and signal recordings $\{\mathbf{x}_i\}$, where the former consists of only noise.
\end{remark}

\begin{remark}\label{re:twoeq}
Consider the eigenvalue decomposition 
$\mathbf{C}_{yy}=\mathbf{U}_y\mathbf{\Sigma}_{yy}\mathbf{U}_y^\top$. 
With $\mathbf{C}_{yy}^{1/2}:=\mathbf{\Sigma}_{yy}^{1/2}\mathbf{U}_{y}^\top$, and the definition 
  $\mathbf{v}:=\mathbf{C}_{yy}^{\top/2}\mathbf{u}\in\mathbb{R}^D$,  \eqref{eq:dpcafm2} can be expressed as	
	\begin{subequations}
		\label{eq:v}
	\begin{align}
		\hat{\mathbf{v}}:=\arg
	\max_{\mathbf{v}\in\mathbb{R}^D}\quad&
		\mathbf{v}^\top\mathbf{C}^{-1/2}_{yy}\mathbf{C}_{xx}\mathbf{C}^{-\top/2}_{yy}\mathbf{v}\\
		{\rm s.\,to}\quad &\mathbf{v}^\top\mathbf{v}=1	
		\end{align}
		\end{subequations}
where $\hat{\mathbf{v}}$ corresponds to the leading eigenvector of $\mathbf{C}_{yy}^{-1/2}\mathbf{C}_{xx} \mathbf{C}_{yy}^{-\top/2}$. Subsequently,  $\hat{\mathbf{u}}$ in \eqref{eq:dpcafm2} is recovered as $\hat{\mathbf{u}}=\mathbf{C}_{yy}^{-\top/2}\hat{\mathbf{v}}$.
This indeed suggests that discriminative analytics of $\{\mathbf{x}_i\}_{i=1}^m$ and $\{\mathbf{y}_j\}_{j=1}^n$ using dPCA  
can be viewed as PCA of the `denoised' or `background-removed' data $\{\mathbf{C}^{-1/2}_{yy}\mathbf{x}_i \}$,
followed by an `inverse' transformation to 
map the obtained subspace vector of the $\{\mathbf{C}^{-1/2}_{yy}\mathbf{x}_i \}$ data  to $\{\mathbf{x}_i\}$ that of the target 
data.  
In this sense, $\{\mathbf{C}^{-1/2}_{yy}\mathbf{x}_i \}$ can be seen as the data obtained after removing the dominant `background' subspace vectors from the target data.

\end{remark}
\begin{remark}
Inexpensive power or Lanczos iterations \cite{saad1} can be employed to compute the principal eigenvectors in \eqref{eq:dpcasol}.
\end{remark}

\begin{algorithm}[t]
	\caption{Discriminative PCA.}
	\label{alg:dpca}
	\begin{algorithmic}[1]
		\STATE {\bfseries Input:}
		Nonzero-mean target and background data $\{\accentset{\circ}{\mathbf{x}}_i\}_{i=1}^m$ and $\{\accentset{\circ}{\mathbf{y}}_j\}_{j=1}^n$; number of dPCs $d$.
		\STATE {\bfseries Exclude} the means from $\{\accentset{\circ}{\mathbf{x}}_i\}$ and $\{\accentset{\circ}{\mathbf{y}}_j\}$ to obtain centered data $\{\mathbf{x}_i\}$, and $\{\mathbf{y}_j\}$. Construct $\mathbf{C}_{xx}$ and $\mathbf{C}_{yy}$.
		\STATE {\bfseries Perform} \label{step:4} eigendecomposition
		on $\mathbf{C}_{yy}^{-1}\mathbf{C}_{xx}$ to obtain the $d$ right eigenvectors $\{\hat{\mathbf{u}}_i\}_{i=1}^d$ associated with the $d$ largest eigenvalues.
		\STATE {\bfseries Output} $\hat{\mathbf{U}}=[\hat{\mathbf{u}}_1\,\cdots \, \hat{\mathbf{u}}_d]$.
		\vspace{-0pt}
	\end{algorithmic}
\end{algorithm}

\textcolor{black}{Consider again \eqref{eq:dpcafm2}. Based on Lagrange duality, when selecting $\alpha=\hat{\lambda}$ in \eqref{eq:cpca}, where $\hat{\lambda}$ is the largest eigenvalue of $\mathbf{C}_{yy}^{-1}\mathbf{C}_{xx}$, cPCA maximizing $\textcolor{black}{{\mathbf{u}}^\top}(\mathbf{C}_{xx}-\hat{\lambda}\mathbf{C}_{yy}){\mathbf{u}}$ is equivalent to $\max_{\mathbf{u}\in\mathbb{R}^{D}}\mathcal{L}(\mathbf{u};\hat{\lambda})=\mathbf{u}^\top(\mathbf{C}_{xx}-\hat{\lambda}\mathbf{C}_{yy})\mathbf{u}+\hat{\lambda}$, which coincides with \eqref{eq:lag} when $\lambda=\hat{\lambda}$ at the optimum. 
This suggests that the optimizers of cPCA and dPCA share the same direction when $\alpha$ in cPCA is chosen to be the optimal dual variable $\hat{\lambda}$ of our dPCA in \eqref{eq:dpcafm2}.
	This equivalence between dPCA and cPCA with a proper $\alpha$ can also be seen from 
	the following.
	\begin{theorem}{\cite[Theorem 2]{guo2003generalized}}
		\label{the:cvsd}
		For real symmetric matrices $\mathbf{C}_{xx}\succeq \mathbf{0}$ and $\mathbf{C}_{yy}\succ\mathbf{0}$, the following holds
			\begin{equation*}
			\check{\lambda}=\frac{\check{\mathbf{u}}^\top\mathbf{C}_{xx}\check{\mathbf{u}}}{\check{\mathbf{u}}^\top\mathbf{C}_{yy}\check{\mathbf{u}}}=\underset{\|\mathbf{u}\|_2=1}{\max}\frac{\mathbf{u}^\top\mathbf{C}_{xx}\mathbf{u}}{\mathbf{u}^\top\mathbf{C}_{yy}\mathbf{u}} 
			\end{equation*}
			if and only if
			\begin{equation*}
			\check{\mathbf{u}}^\top(\mathbf{C}_{xx}-\check{\lambda}\mathbf{C}_{yy})\check{\mathbf{u}}=\underset{\|\mathbf{u}\|_2=1}{\max}\mathbf{u}^\top(\mathbf{C}_{xx}-\check{\lambda}\mathbf{C}_{yy})\mathbf{u}.
			\end{equation*}
	\end{theorem}
	}

To gain further insight into the relationship between  dPCA and cPCA,
suppose that  $\mathbf{C}_{xx}$ and $\mathbf{C}_{yy}$ are simultaneously diagonalizable; that is, there exists an unitary matrix $\mathbf{U}\in\mathbb{R}^{D\times D}$ such that
\begin{equation*}
\mathbf{C}_{xx}:=\mathbf{U}\mathbf{\Sigma}_{xx}\mathbf{U}^\top,\quad {\rm and}\quad \mathbf{C}_{yy}:=\mathbf{U}\mathbf{\Sigma}_{yy}\mathbf{U}^\top
\end{equation*} 
where diagonal matrices $\mathbf{\Sigma}_{xx},\,\mathbf{\Sigma}_{yy}\succ \mathbf{0}$ hold accordingly eigenvalues $\{\lambda_x^i\}_{i=1}^D$ of $\mathbf{C}_{xx}$ and $\{\lambda_y^i\}_{i=1}^D$ of $\mathbf{C}_{yy}$ on their main diagonals. 
\textcolor{black}{Even if the two datasets may share some subspace vectors, $\{\lambda_x^i\}_{i=1}^D$ and $\{\lambda_y^i\}_{i=1}^D$ are in general not the same.}
It is easy to check that $\mathbf{C}_{yy}^{-1}\mathbf{C}_{xx}=\mathbf{U}\mathbf{\Sigma}_{yy}^{-1}\mathbf{\Sigma}_{xx}\mathbf{U}^\top=\mathbf{U} {\rm diag}\big(\{\frac{\lambda_x^i}{\lambda_y^i}\}_{i=1}^D\big)\mathbf{U}^\top$. Seeking the first $d$ latent subspace vectors is tantamount to taking the $d$ columns of $\mathbf{U}$ that correspond to the $d$ largest values among $\{\frac{\lambda_x^i}{\lambda_y^i}\}_{i=1}^D$.
On the other hand, cPCA for a fixed $\alpha$, looks for the first $d$ latent subspace vectors of $\mathbf{C}_{xx}-\alpha\mathbf{C}_{yy}=\mathbf{U}(\mathbf{\Sigma}_{xx}-\alpha{\bm\Sigma}_{yy})\mathbf{U}^\top=\mathbf{U}{\rm diag}\big(\{\lambda_x^i-\alpha\lambda_y^i\}_{i=1}^D\big)\mathbf{U}^\top$, which amounts to taking the $d$ columns of $\mathbf{U}$ associated with the $d$ largest values in $\{\lambda_x^i-\alpha\lambda_y^i\}_{i=1}^D$. 
\textcolor{black}{This further confirms that when $\alpha$ is sufficiently large (small), cPCA returns the $d$ columns of $\mathbf{U}$ associated with the $d$  largest $\lambda_{y}^{i}$'s ($\lambda_x^{i}$'s). When $\alpha$ is not properly chosen, cPCA may fail to extract the most contrastive information from target data relative to background data. In contrast, this is not an issue is not present in dPCA simply because it has no tunable parameter.}

\section{Optimality of {d}PCA}
\label{sec:optim}

In this section, we show that dPCA is optimal when data obey a certain affine model. In a similar vein, PCA adopts  a factor analysis model  to express the non-centered background data $\{\accentset{\circ}{\mathbf{y}}_j\in\mathbb{R}^D\}_{j=1}^n$ as
\begin{equation}
\accentset{\circ}{\mathbf{y}}_j=\mathbf{m}_y+
\mathbf{U}_b\bm{\psi}_j+\mathbf{e}_{y,j},	\quad j=1,\,\ldots,\,n\label{eq:y}
\end{equation}
where $\mathbf{m}_y\in\mathbb{R}^D$ denotes the unknown location (mean) vector; $\mathbf{U}_b\in\mathbb{R}^{D\times k}$ has orthonormal columns with $k<D$;
 $\{\bm{\psi}_j\in\mathbb{R}^k\}_{j=1}^n$ are some unknown coefficients with covariance matrix $\bm{\Sigma}_b:={\rm diag}(\lambda_{y,1},\,\lambda_{y,2},\,\ldots,\,\lambda_{y,k})\in\mathbb{R}^{k\times k}$;
 and the modeling errors $\{\mathbf{e}_{y,j}\in\mathbb{R}^D\}_{j=1}^n$ are assumed  zero-mean   with covariance matrix $\mathbb{E}[\mathbf{e}_{y,j}\mathbf{e}_{y,j}^\top]=\mathbf{I}$. Adopting the least-squares (LS) criterion,
 the unknowns $\mathbf{m}_y$, $\mathbf{U}_b$, and $\{\bm{\psi}_j\}$ can be estimated by \cite{1995pca} 
 \begin{equation*}
\underset{\mathbf{m}_y,\{\bm{\psi}_j\}\atop
\mathbf{U}_b
}{\min}\quad \sum_{j=1}^n\left\|\accentset{\circ}{\mathbf{y}}_j-\mathbf{m}_y-\mathbf{U}_b\bm{\psi}_j\right\|_2^2\quad{\rm s.\,to}\quad  \mathbf{U}_b^\top\mathbf{U}_b=\mathbf{I}
\end{equation*}
we find at the optimum 
 $\hat{\mathbf{m}}_y:=(1/n)\sum_{j=1}^n{\accentset{\circ}{\mathbf{y}}}_j$, $\big\{\hat{\bm{\psi}}_j:=\hat{\mathbf{U}}_b^\top(\accentset{\circ}{\mathbf{y}}_j-\hat{\mathbf{m}}_y)\big\}$, with $\hat{\mathbf{U}}_b$ columns given by the first $k$ leading eigenvectors of
$\mathbf{C}_{yy}=(1/n)\sum_{j=1}^n\mathbf{y}_j\mathbf{y}_j^\top$, 
in which $\mathbf{y}_j:={\accentset{\circ}{\mathbf{y}}}_j-\hat{\mathbf{m}}_y$.  
It is clear that $\mathbb{E}[\mathbf{y}_j\mathbf{y}_j^\top]=\mathbf{U}_b\bm{\Sigma}_{b}\mathbf{U}_b^\top+\mathbf{I}$.
Let matrix $\mathbf{U}_n\in\mathbb{R}^{D\times (D-k)}$ with orthonormal columns satisfying $\mathbf{U}_n^\top\mathbf{U}_b=\mathbf{0}$, and $\mathbf{U}_y:=[\mathbf{U}_b~\mathbf{U}_n]\in\mathbb{R}^{D\times D}$ with $\bm{\Sigma}_y:={\rm diag}(\{\lambda_{y,i}\}_{i=1}^D)$, where $\{\lambda_{y,k+\ell}:=1\}_{\ell=1}^{D-k}$. Therefore, $\mathbf{U}_b\bm{\Sigma}_{b}\mathbf{U}_b^\top+\mathbf{I}=\mathbf{U}_y\bm{\Sigma}_y\mathbf{U}_y^\top$. As $n\to\infty$, the strong law of large numbers asserts that $\mathbf{C}_{yy}\to\mathbb{E}[\mathbf{y}_j\mathbf{y}_j^\top]$; that is,  $\mathbf{C}_{yy}=\mathbf{U}_y\bm{\Sigma}_y\mathbf{U}_y^\top$ as $n\to \infty$.

Here we assume that the target data  $\{\accentset{\circ}{\mathbf{x}}_i\in\mathbb{R}^D\}_{i=1}^m$ share the  background related matrix $\mathbf{U}_b$ with data $\{\accentset{\circ}{\mathbf{y}}_j\}$, but also have $d$ extra  vectors 
 specific to the target data relative to the background data. 
 \textcolor{black}{This assumption is well justified in realistic setups. In the example discussed in Sec. \ref{sec:intro}, both patients'  and healthy persons' gene-expression data contain  common patterns corresponding to geographical and gender variances; while the patients' gene-expression data contain some specific latent subspace vectors corresponding to their diseases.}
 Focusing for simplicity  on $d=1$, we model $\{\accentset{\circ}{\mathbf{x}}_i\}$ as
\begin{equation}
\accentset{\circ}{\mathbf{x}}_i=\mathbf{m}_x+\left[\mathbf{U}_b~\;\mathbf{u}_s\right]\left[\!\!\begin{array}{c}\boldsymbol{\chi}_{b,i}\\{\chi}_{s,i}\end{array}\!\!\right]+\mathbf{e}_{x,i},\quad i=1,\,\ldots,\,m\label{eq:x}
\end{equation}
where $\mathbf{m}_x\in\mathbb{R}^D$ represents the location of $\{\accentset{\circ}{\mathbf{x}}_i\}_{i=1}^m$; $\{\mathbf{e}_{x,i}\}_{i=1}^m$ account for zero-mean modeling errors;  $\mathbf{U}_x:=[\mathbf{U}_b~\mathbf{u}_s]\in\mathbb{R}^{D\times (k+1)}$ collects orthonormal columns, where $\mathbf{U}_b$ is the shared latent subspace vectors associated 
 with background data, and $\mathbf{u}_s\in\mathbb{R}^{D}$ is 
\textcolor{black}{a latent subspace vector
	 unique to the target data, but not to the background data.}
Simply put, 
our goal is to extract this discriminative subspace $\mathbf{u}_s$ given $\{\accentset{\circ}{\mathbf{x}}_i\}_{i=1}^m$ and $\{\accentset{\circ}{\mathbf{y}}_j\}_{j=1}^n$.

Similarly, given $\{\accentset{\circ}{\mathbf{x}}_i\}$, the unknowns $\mathbf{m}_x$, $\mathbf{U}_x$, and
 $\{\bm{\chi}_i:=[\bm{\chi}_{b,i}^\top, \,\chi_{s,i}]^\top\}$
 can be estimated by 
  	\begin{equation*}
 	\underset{\mathbf{m}_x,\,\{\bm{\chi}_i\}\atop\mathbf{U}_x}{\max}
 	\quad\sum_{i=1}^m\left\|\accentset{\circ}{\mathbf{x}}_i-\mathbf{m}_x-\mathbf{U}_x\bm{\chi}_i\right\|_2^2
	\quad{\rm s.\,to}\quad \mathbf{U}_x^\top\mathbf{U}_x=\mathbf{I} 
 	\end{equation*}
 	yielding $\hat{\mathbf{m}}_x:=(1/m)\sum_{i=1}^{m}\accentset{\circ}{\mathbf{x}}_i$, $\hat{\bm{\chi}}_i:=\hat{\mathbf{U}}_x^\top\mathbf{x}_i$ with $\mathbf{x}_i:=\accentset{\circ}{\mathbf{x}}_i-\hat{\mathbf{m}}_x$,
 where $\hat{\mathbf{U}}_x$ has columns the \textcolor{black}{$(k+1)$ principal} eigenvectors of $\mathbf{C}_{xx}=(1/m)\sum_{i=1}^{m}\mathbf{x}_i\mathbf{x}_i^\top$.
When $m\to \infty$, it holds that $\mathbf{C}_{xx}
=\mathbf{U}_x \bm{\Sigma}_x \mathbf{U}_x^\top$, with $\bm{\Sigma}_x:=\mathbb{E}[\bm{\chi}_i\bm{\chi}_i^\top]={\rm diag}(\lambda_{x,1},\,\lambda_{x,2},\,\ldots,\,\lambda_{x,k+1})\in\mathbb{R}^{(k+1)\times (k+1)}$.

Let $\bm{\Sigma}_{x,k}\in\mathbb{R}^{k\times k}$ denote the submatrix of $\bm{\Sigma}_x$ formed by its first $k$ rows and columns. When $m,\,n\to\infty$ and $\mathbf{C}_{yy}$ is nonsingular, one can express $\mathbf{C}_{yy}^{-1}\mathbf{C}_{xx}$ as  
\begin{align*}
&\mathbf{U}_y\bm{\Sigma}_y^{-1}\mathbf{U}_y^\top \mathbf{U}_x\bm{\Sigma}_x\mathbf{U}_x^\top\\
&=[\mathbf{U}_b ~ \mathbf{U}_n]
\left[
\!\begin{array}
{cc}
\bm{\Sigma}_{b}^{-1} &\mathbf{0}\\
\mathbf{0}  &\mathbf{I}
\end{array}\!
\right]
\left[\!
\begin{array}
{cc}
\mathbf{I}&\mathbf{0}\\
\mathbf{0}&\mathbf{U}_n^\top \mathbf{u}_s
\end{array}
\!
\right]\\
&\quad \times
\left[
\!\begin{array}
{cc}
\bm{\Sigma}_{x,k}&\mathbf{0}\\
\mathbf{0}&\lambda_{x,k+1}
\end{array}
\!
\right]
\left[
\!
\begin{array}
{c}
\mathbf{U}_b^\top\\
\mathbf{u}_s^\top
\end{array}
\!
\right]\\
&=[\mathbf{U}_b ~ \mathbf{U}_n]
\left[\!
\begin{array}
{cc}
\bm{\Sigma}_{b}^{-1}\bm{\Sigma}_{x,k}&\mathbf{0}\\
\mathbf{0}& \lambda_{x,k+1}\mathbf{U}_n^\top \mathbf{u}_s
\end{array}
\!
\right]
\left[
\!
\begin{array}
{c}
\mathbf{U}_b^\top\\
\mathbf{u}_s^\top
\end{array}
\!
\right]\\
&=\mathbf{U}_b \bm{\Sigma}_{b}^{-1}\bm{\Sigma}_{x,k} \mathbf{U}_b^\top + \lambda_{x,k+1}\mathbf{U}_n\mathbf{U}_n^\top \mathbf{u}_s\mathbf{u}_s^\top.
\end{align*}
Observe that the first and second summands 
have rank $k$ and $1$, respectively,
  implying that $\mathbf{C}_{yy}^{-1}\mathbf{C}_{xx}$ has at most  rank $k+1$. If $\mathbf{u}_{b,i}$ denotes the $i$-th column of $\mathbf{U}_b$, that is orthogonal to $\{\mathbf{u}_{b,j}\}_{j=1,j\ne i}^k$ and $\mathbf{u}_s$, right-multiplying $\mathbf{C}_{yy}^{-1}\mathbf{C}_{xx}$ by $\mathbf{u}_{b,i}$ yields
\begin{equation*}
	\mathbf{C}_{yy}^{-1}\mathbf{C}_{xx}\mathbf{u}_{b,i} = ({\lambda_{x,i}}/{\lambda_{y,i}})\,\mathbf{u}_{b,i}
\end{equation*}
for $i=1,\,\ldots,\,k$, which hints that
 $\{\mathbf{u}_{b,i}\}_{i=1}^k$ are $k$ eigenvectors of $\mathbf{C}_{yy}^{-1}\mathbf{C}_{xx}$ associated with eigenvalues $\{{\lambda_{x,i}}/{\lambda_{y,i}}\}_{i=1}^k$. 
 Again, right-multiplying $\mathbf{C}_{yy}^{-1}\mathbf{C}_{xx}$ by $\mathbf{u}_s$ gives rise to
\begin{equation}\label{eq:yxus}
	\mathbf{C}_{yy}^{-1}\mathbf{C}_{xx}\mathbf{u}_s =\lambda_{x,k+1} \mathbf{U}_n \mathbf{U}_n^\top \mathbf{u}_s\mathbf{u}_s^\top \mathbf{u}_s = \lambda_{x,k+1} \mathbf{U}_n \mathbf{U}_n^\top \mathbf{u}_s.
\end{equation}
To proceed, we will leverage  the following three facts: i) $\mathbf{u}_s$ is orthogonal to all columns of $\mathbf{U}_b$; ii) columns of $\mathbf{U}_n$ are orthogonal to those of $\mathbf{U}_b$; and iii) $[\mathbf{U}_b~\mathbf{U}_n]$ has full rank. Based on i)-iii), 
 it follows readily that $\mathbf{u}_s$ can be uniquely expressed as a linear combination of columns of $\mathbf{U}_n$; that is, $\mathbf{u}_s:=\sum_{i=1}^{D-k}p_i\mathbf{u}_{n,i}$, where $\{p_i\}_{i=1}^{D-k}$ are some unknown coefficients, and $\mathbf{u}_{n,i}$ denotes the $i$-th column of $\mathbf{U}_n$. One can manipulate  $\mathbf{U}_n \mathbf{U}_n^\top \mathbf{u}_s$ in \eqref{eq:yxus} as 
\begin{align*}
	\mathbf{U}_n \mathbf{U}_n^\top \mathbf{u}_s&=[\mathbf{u}_{n,1}~\cdots~\mathbf{u}_{n,D-k}]
	\left[\!
	\begin{array}{c}
	\mathbf{u}_{n,1}^\top \mathbf{u}_s\\
    \vdots\\
	\mathbf{u}_{n,D-k}^\top \mathbf{u}_s
	\end{array}\!
	\right]\\
	&=\mathbf{u}_{n,1}\mathbf{u}_{n,1}^\top\mathbf{u}_s+\,\cdots\,+\mathbf{u}_{n,D-k}\mathbf{u}_{n,D-k}^\top\mathbf{u}_s\\
	&=p_1\mathbf{u}_{n,1}+\,\cdots\,+p_{D-k}\mathbf{u}_{n,D-k}\\
	&=\mathbf{u}_s
\end{align*}
yielding $\mathbf{C}_{yy}^{-1}\mathbf{C}_{xx}\mathbf{u}_s=\lambda_{x,k+1}\mathbf{u}_s$; that is, $\mathbf{u}_s$ is the $(k+1)$-st eigenvector of $\mathbf{C}_{yy}^{-1}\mathbf{C}_{xx}$ corresponding to eigenvalue $\lambda_{x,k+1}$.

Before moving on, we will make two assumptions.

\begin{assumption}\label{asmp:model}
	Background and target data  are generated according to the models \eqref{eq:y} and \eqref{eq:x}, respectively, with the background data sample covariance matrix being nonsingular.
\end{assumption}
\begin{assumption}\label{asmp:unique}
	It holds for all $i=1,\,\ldots,\,k$ that ${\lambda_{x,k+1}}/{\lambda_{y,k+1}}>{\lambda_{x,i}}/{\lambda_{y,i}}$.
\end{assumption}
Assumption \ref{asmp:unique} essentially requires that $\mathbf{u}_s$ is discriminative enough in the target data relative to the background data. 
\textcolor{black}{After combining Assumption \ref{asmp:unique} and the fact that $\mathbf{u}_s$ is an eigenvector of $\mathbf{C}_{yy}^{-1}\mathbf{C}_{xx}$, it follows readily that the eigenvector of $\mathbf{C}_{yy}^{-1}\mathbf{C}_{xx}$ associated with the largest eigenvalue is $\mathbf{u}_s$.}
Under these two assumptions, we establish the optimality of dPCA next. 
\begin{theorem}
	\label{the:optm}
	Under Assumptions \ref{asmp:model} and \ref{asmp:unique} with $d=1$, as $m,\,n\to\infty$, the solution of \eqref{eq:dpca} recovers the subspace vector specific to target data relative to background data, namely $\mathbf{u}_s$. 
\end{theorem}

\section{Kernel dPCA} \label{sec:kdpca}

With advances in data acquisition and data storage technologies, a sheer volume of possibly high-dimensional data are collected daily, that topologically lie on a  nonlinear manifold in general. This goes beyond the ability of the (linear) dPCA in Sec. \ref{sec:dpca} due mainly to a couple of reasons: i) dPCA presumes a linear low-dimensional hyperplane to project the target data vectors; and ii) dPCA incurs computational complexity  $\mathcal{O}(\max (m,\,n) D^2)$ that grows quadratically with the dimensionality of data vectors.
 To address these challenges, this section generalizes dPCA to account for nonlinear data relationships via kernel-based learning, and puts forth kernel (K) dPCA for nonlinear discriminative analytics. Specifically, KdPCA starts by \textcolor{black}{mapping} both the target and background data vectors from the original data space to a higher-dimensional (possibly infinite-dimensional) feature space using a common nonlinear function, which is followed by performing linear dPCA on the \textcolor{black}{transformed} data.

Consider first  the dual version of dPCA, starting with  the $N:=m+n$ augmented data  $\{\mathbf{z}_i\in\mathbb{R}^D\}_{i=1}^N$ as 
\begin{equation*}
\label{eq:z}
\mathbf{z}_i:=\left\{
\begin{array}{ll}
\mathbf{x}_i, & 1\le i \le m\\
\mathbf{y}_{i-m}, & m< i \le N
\end{array}
\right.
\end{equation*}
and express the wanted
subspace vector $\mathbf{u}\in\mathbb{R}^D$ in terms of
 $\mathbf{Z}:=[\mathbf{z}_1\,\cdots \, \mathbf{z}_N]\in\mathbb{R}^{D\times N}$, yielding $\mathbf{u}:=\mathbf{Z}\mathbf{a}$, where $\mathbf{a}\in\mathbb{R}^N$ denotes  the dual vector. \textcolor{black}{When ${\rm min}(m\,,n)\gg D$, matrix $\mathbf{Z}$ has full row rank in general. Thus, there always exists a vector $\mathbf{a}$ so that $\mathbf{u}=\mathbf{Z}\mathbf{a}$. Similar steps have also been used in obtaining dual versions of PCA and CCA \cite{kpca,2004kernel}.} 
Substituting $\mathbf{u}=\mathbf{Z}\mathbf{a}$ into \eqref{eq:dpca} leads to our dual dPCA 
\begin{equation}
\textcolor{black}{	\label{eq:ddpca}
	\underset{\mathbf{a}\in\mathbb{R}^N}{\max}
	\quad \frac{\mathbf{a}^\top \mathbf{Z}^\top \mathbf{C}_{xx}\mathbf{Z}\mathbf{a}}{ \mathbf{a}^\top \mathbf{Z}^\top\mathbf{C}_{yy}\mathbf{Z}\mathbf{a}}
}
\end{equation}
based on which we will develop our KdPCA in the sequel.

Similar to deriving KPCA from dual PCA \cite{kpca}, our approach is first to transform $\{\mathbf{z}_i\}_{i=1}^N$ from $\mathbb{R}^D$ to a high-dimensional space $\mathbb{R}^L$ (possibly with $L=\infty$) by some nonlinear mapping function $\bm{\phi}(\cdot)$, followed by removing the sample means of $\{\bm{\phi}(\mathbf{x}_i)\}$ and $\{\bm{\phi}(\mathbf{y}_j)\}$ from the corresponding transformed data; and subsequently, implementing dPCA on the centered transformed datasets to obtain the low-dimensional \textcolor{black}{kernel} dPCs. Specifically,  the sample covariance matrices of  $\{\bm{\phi}(\mathbf{x}_i)\}_{i=1}^m$ and $\{\bm{\phi}(\mathbf{y}_j)\}_{j=1}^n$ can be expressed as 
\begin{align*}	
	\mathbf{C}_{xx}^{\phi }&:=\frac{1}{m}\sum_{i=1}^m \left(\bm{\phi}(\mathbf{x}_i)-\bm{\mu}_{ x}\right)\left(\bm{\phi}(\mathbf{x}_i)-\bm{\mu}_{ x}\right)^\top\in\mathbb{R}^{L\times L}\\
	\mathbf{C}_{yy}^{\phi }&:=\frac{1}{n}\sum_{j=1}^n \left(\bm{\phi}(\mathbf{y}_j)-\bm{\mu}_{ y}\right)\left(\bm{\phi}(\mathbf{y}_j)-\bm{\mu}_{ y}\right)^\top\in\mathbb{R}^{L\times L}
\end{align*}
where the $L$-dimensional vectors $\bm{\mu}_{x}:=(1/m)\sum_{i=1}^m\bm{\phi}(\mathbf{x}_i)
$ and $\bm{\mu}_{y}:=(1/n)\sum_{j=1}^n\bm{\phi}(\mathbf{y}_j)
$ are accordingly the sample means of $\{\bm{\phi}(\mathbf{x}_i)\}$ and $\{\bm{\phi}(\mathbf{y}_j)\}$.
For convenience, let
 $\bm{\Phi}(\mathbf{Z}):=[\bm{\phi}(\mathbf{x}_1)-\bm{\mu}_x,\,\cdots,\,\bm{\phi}(\mathbf{x}_m)-\bm{\mu}_x,\,\bm{\phi}(\mathbf{y}_1)-\bm{\mu}_y,\,\cdots,\,\bm{\phi}(\textcolor{black}{\mathbf{y}_n})-\bm{\mu}_y]\in\mathbb{R}^{L\times N}$.
  Upon replacing $\{\mathbf{x}_i\}$ and $\{\mathbf{y}_j\}$ in \eqref{eq:ddpca} with $\{\bm{\phi}(\mathbf{x}_i)-\bm{\mu}_x\}$ and $\{\bm{\phi}(\mathbf{y}_j)-\bm{\mu}_y\}$, respectively,  \textcolor{black}{the kernel version of \eqref{eq:ddpca} boils down to}
\textcolor{black}{
\begin{equation}	
	\label{eq:kdpca}
	\underset{\mathbf{a}\in\mathbb{R}^N}{\max}
	\quad\frac{\mathbf{a}^\top \bm{\Phi}^\top(\mathbf{Z}) \mathbf{C}_{xx}^{\phi}\bm{\Phi}(\mathbf{Z})\mathbf{a}}{ \mathbf{a}^\top \bm{\Phi}^\top(\mathbf{Z})\mathbf{C}_{y y}^{\phi}\bm{\Phi}(\mathbf{Z})\mathbf{a}}.
\end{equation}}

In the sequel, \eqref{eq:kdpca} will be further simplified by leveraging the so-termed `kernel trick' \cite{RKHS}. 

To start, define a kernel matrix $\mathbf{K}_{xx}\in\mathbb{R}^{m\times m}$ of  $\{\mathbf{x}_i\}$ whose $(i,\,j)$-th entry is $\kappa(\mathbf{x}_i,\,\mathbf{x}_j):=\left<\bm{\phi}(\mathbf{x}_i),\,\bm{\phi}(\mathbf{x}_j)\right>$ for $i,\,j=1,\,\ldots,\,m$, where $\kappa(\cdot)$ represents some kernel function. Matrix $\mathbf{K}_{yy}\in\mathbb{R}^{n\times n}$ of $\{\mathbf{y}_j\}$ is defined likewise. Further, the $(i,\,j)$-th entry of matrix $\mathbf{K}_{xy}\in\mathbb{R}^{m\times n}$ is  $\kappa(\mathbf{x}_i,\,\mathbf{y}_j):=\left<\bm{\phi}(\mathbf{x}_i),\,\bm{\phi}(\mathbf{y}_j)\right>$. Centering $\mathbf{K}_{xx}$, $\mathbf{K}_{yy}$, and $\mathbf{K}_{xy}$ produces 
\begin{align*}
\mathbf{K}_{xx}^c&:=\mathbf{K}_{xx}-\tfrac{1}{m}\mathbf{1}_{m }\mathbf{K}_{xx}-\tfrac{1}{m}\mathbf{K}_{xx}\mathbf{1}_{m}+\tfrac{1}{m^2}\mathbf{1}_{m}\mathbf{K}_{xx}\mathbf{1}_{m}\\
\mathbf{K}_{yy}^c&:=\mathbf{K}_{yy}-\!\tfrac{1}{n}\mathbf{1}_{ n}\mathbf{K}_{yy}-\!\tfrac{1}{n}\mathbf{K}_{yy}\mathbf{1}_{n}+\!\tfrac{1}{n^2}\mathbf{1}_{n}\mathbf{K}_{yy}\mathbf{1}_{n}\\
\mathbf{K}_{xy}^c&:=\mathbf{K}_{xy}-\tfrac{1}{m}\mathbf{1}_{ m}\mathbf{K}_{xy}-\tfrac{1}{n}\mathbf{K}_{xy}\mathbf{1}_{ n}+\tfrac{1}{mn}\mathbf{1}_{m}\mathbf{K}_{xy}\mathbf{1}_{n}
\end{align*}
with matrices $\mathbf{1}_{m}\in\mathbb{R}^{m\times m}$ and $\mathbf{1}_n\in\mathbb{R}^{n\times n}$ having all entries $1$. Based on those centered matrices, let
\begin{equation}
\label{eq:k}
\mathbf{K}:=\left[\begin{array}
{cc}
\mathbf{K}_{xx}^c &\mathbf{K}_{xy}^c\\
(\mathbf{K}_{xy}^c)^\top  &\mathbf{K}_{yy}^c
\end{array}
\right]\in\mathbb{R}^{N\times N}.
\end{equation}
Define further  
$\mathbf{K}^x\in\mathbb{R}^{N\times N}$ and $\mathbf{K}^y\in\mathbb{R}^{N\times N}$ with $(i,\,j)$-th entries
\begin{subequations}
	\label{eq:kxky}
	\begin{align}
	K^x_{i,j}\,:=\left\{
	\begin{array}{ll}
	K_{i,j}/m \!&1\le i \le m\\
	~~~0 &m<i\le N
	\end{array}
	\right.\label{eq:kx}\\
	K^y_{i,j}\,:=\left\{
	\begin{array}{ll}
	~~~0 \!&1\le i \le m\\
	K_{i,j}/n &m<i\le N
	\end{array}
	\right.\label{eq:ky}
	\end{align}
\end{subequations}
where $K_{i,\,j}$ stands for the $(i,\,j)$-th entry of $\mathbf{K}$.

\begin{algorithm}[t]
	\caption{Kernel dPCA.}
	\label{alg:kdpca}
	\begin{algorithmic}[1]
		\STATE {\bfseries Input:}
		Target data $\{\mathbf{x}_i\}_{i=1}^m$ and background data $\{\mathbf{y}_j\}_{j=1}^n$; number of dPCs $d$; kernel function $\kappa(\cdot)$; constant $\epsilon$.
		\STATE {\bfseries Construct} $\mathbf{K}$ using \eqref{eq:k}. Build $\mathbf{K}^x$ and $\mathbf{K}^y$ via \eqref{eq:kxky}.
		\STATE {\bfseries Solve} \eqref{eq:kdpcasol} to obtain the first $d$  eigenvectors $\{\hat{\mathbf{a}}_i\}_{i=1}^d$.
		\STATE {\bfseries Output} $\hat{\mathbf{A}}:=[\hat{\mathbf{a}}_1\,\cdots\,\hat{\mathbf{a}}_d]$.
		\vspace{-0pt}
	\end{algorithmic}
\end{algorithm}

\textcolor{black}{Using \eqref{eq:k} and \eqref{eq:kxky}, it can be easily verified that $\mathbf{K}\mathbf{K}^x=\mathbf{K}{\rm diag}(\{\iota_i^x\}_{i=1}^N)\mathbf{K}$, and $\mathbf{K}\mathbf{K}^y=\mathbf{K}{\rm diag}(\{\iota_i^y\}_{i=1}^N)\mathbf{K}$, where $\{\iota_i^x=1\}_{i=1}^m$, $\{\iota_i^x=0\}_{i=m+1}^N$, $\{\iota_i^y=0\}_{i=1}^m$, and $\{\iota_i^y=1\}_{i=m+1}^N$. That is, both  $\mathbf{K}\mathbf{K}^x$ and $\mathbf{K}\mathbf{K}^y$ are symmetric. }
Substituting \eqref{eq:k} and \eqref{eq:kxky} into \eqref{eq:kdpca} yields \textcolor{black}{(see details in the Appendix)}
\begin{equation}	\label{eq:kdpcafm2o}
%	\hat{\mathbf{a}}:=\arg	
	\underset{\mathbf{a}\in\mathbb{R}^N}{\max}
	\quad\frac{\mathbf{a}^\top \mathbf{K}\mathbf{K}^x\mathbf{a}}{ \mathbf{a}^\top \mathbf{K}\mathbf{K}^y\mathbf{a}}.
\end{equation} 
\textcolor{black}{Due to the rank-deficiency of $\mathbf{K}\mathbf{K}^y$ however, \eqref{eq:kdpcafm2o} does not admit a meaningful solution. 
	To address this issue, following kPCA \cite{kpca,2004kernel}, 
	a positive constant $\epsilon>0$ is added to the diagonal entries of $\mathbf{K}\mathbf{K}^y$. Hence, our KdPCA formulation for $d=1$ is given by}
	\begin{equation}	\label{eq:kdpcafm2}
\textcolor{black}{		\hat{\mathbf{a}}:=\arg	
	\underset{\mathbf{a}\in\mathbb{R}^N}{\max}
	\quad\frac{\mathbf{a}^\top \mathbf{K}\mathbf{K}^x\mathbf{a}}{ \mathbf{a}^\top \left(\mathbf{K}\mathbf{K}^y+\epsilon \mathbf{I}\right)\mathbf{a}}.}
	\end{equation}
Along the lines of  dPCA, the solution of KdPCA in \eqref{eq:kdpcafm2} can be provided by 
\begin{equation}
\label{eq:kdpcasol}
\textcolor{black}{
\left(\mathbf{K}\mathbf{K}^y+\epsilon\mathbf{I}\right)^{-1}\mathbf{K}\mathbf{K}^x\hat{\mathbf{a}}=\hat{\lambda}
\hat{\mathbf{a}}.}
\end{equation}
The optimizer $\hat{\mathbf{a}}$ coincides with the right eigenvector of $\left(\mathbf{K}\mathbf{K}^y+\epsilon\mathbf{I}\right)^{-1}\mathbf{K}\mathbf{K}^x$ corresponding to the largest eigenvalue $\hat{\lambda}=\lambda_1$.

 When looking for $d$ dPCs, with $\{\mathbf{a}_i\}_{i=1}^d$ collected as columns in $\mathbf{A}:=[\mathbf{a}_1\,\cdots \,\mathbf{a}_d]\in\mathbb{R}^{N\times d}$, the KdPCA in  \eqref{eq:kdpcafm2} can be generalized to $d\ge 2$ as 
	\begin{equation*}
			\hat{\mathbf{A}}:=\arg
		\underset{\mathbf{A}\in\mathbb{R}^{N\times d}}{\max}
		\quad {\rm Tr} \left[ \left(\mathbf{A}^\top (\mathbf{K}\mathbf{K}^y+\epsilon \mathbf{I}\right)\mathbf{A})^{-1} \mathbf{A}^\top \mathbf{K}\mathbf{K}^x\mathbf{A}\right]
	\end{equation*}
whose columns correspond to the $d$ right eigenvectors of $\left(\mathbf{K}\mathbf{K}^y+\epsilon\mathbf{I}\right)^{-1}\mathbf{K}\mathbf{K}^x$
 associated with the $d$ largest  eigenvalues.
   Having found  $\hat{\mathbf{A}}$, one can project the data $\bm{\Phi}(\mathbf{Z})$ onto the obtained $d$ subspace vectors by $\mathbf{K}\hat{\mathbf{A}}$. 
 It is worth remarking that KdPCA can be performed in the high-dimensional feature space without explicitly forming and evaluating the nonlinear transformations. Indeed, this becomes possible by the `kernel trick' \cite{RKHS}.   
 The main steps of KdPCA are given in Alg. \ref{alg:kdpca}. 
 
Two remarks are worth making at this point.
 	\begin{remark}
 		When the kernel function required to form $\mathbf{K}_{xx}$, $\mathbf{K}_{yy}$, and $\mathbf{K}_{xy}$ is not given, one may use the multi-kernel learning method to automatically choose the right kernel function(s); see for example, \cite{mkl2004,zhang2017going,tsp2017zwrg}. Specifically, one can presume $\mathbf{K}_{xx}:=\sum_{i=1}^P\delta_i\mathbf{K}_{xx}^i$, $\mathbf{K}_{yy}:=\sum_{i=1}^P\delta_i\mathbf{K}_{yy}^i$, and $\mathbf{K}_{xy}:=\sum_{i=1}^P\delta_i\mathbf{K}_{xy}^i$ in \eqref{eq:kdpcafm2}, where $\mathbf{K}_{xx}^i\in\mathbb{R}^{m\times m}$, $\mathbf{K}_{yy}^i\in\mathbb{R}^{n\times n}$, and $\mathbf{K}_{xy}^i\in\mathbb{R}^{m\times n}$ are formed using the kernel function $\kappa_i(\cdot)$; and $\{\kappa_i(\cdot)\}_{i=1}^P$ are a preselected dictionary of known kernels, but  $\{\delta_i\}_{i=1}^P$ will be treated as unknowns to be learned along with $\mathbf{A}$ in \eqref{eq:kdpcafm2}.
 	\end{remark}
 	\begin{remark}
 		In the absence of  background data, upon setting $\{\bm{\phi}(\mathbf{y}_j)=\mathbf{0}\}$, and $\epsilon=1$ in \eqref{eq:kdpcafm2}, matrix $\left(\mathbf{K}\mathbf{K}^y+\epsilon\mathbf{I}\right)^{-1}\mathbf{K}\mathbf{K}^x$ reduces to
 		\begin{equation*}
 		\mathbf{M}:=\left[
 		\begin{array}
 		{cc}
 		(\mathbf{K}_{xx}^c)^2 &\mathbf{0}\\
 		\mathbf{0}  &\mathbf{0}
 		\end{array}
 		\right].
 		\end{equation*} After collecting the first $m$ entries of $\hat{\mathbf{a}}_i$ into $\mathbf{w}_i\in\mathbb{R}^{m}$, \eqref{eq:kdpcasol} suggests that $(\mathbf{K}_{xx}^c)^2\mathbf{w}_i=\lambda_i\mathbf{w}_i$, where $\lambda_i$ denotes the $i$-th largest eigenvalue of $\mathbf{M}$. 
 		Clearly, $\{\mathbf{w}_i\}_{i=1}^d$ can be viewed as the $d$ eigenvectors of $(\mathbf{K}_{xx}^c)^2$  associated with their $d$ largest eigenvalues. Recall that KPCA finds the first $d$ principal eigenvectors of $\mathbf{K}_{xx}^c$ \cite{kpca}. Thus, KPCA is a special case of KdPCA, when no background data are employed.
 	\end{remark}

\section{Discriminative Analytics with\\ Multiple Background Datasets} \label{sec:mdpca}

So far, we have presented discriminative analytics methods for two datasets. 
This section presents their generalizations
to cope with multiple (specifically, one target plus more than one background) datasets.  
Suppose that, in addition to the zero-mean target dataset $\{\mathbf{x}_i\in\mathbb{R}^D\}_{i=1}^m$, we are also given $M\ge 2$ centered background datasets $\{\mathbf{y}_j^k\}_{j=1}^{n_k}$ for $k=1,\,\ldots,\,M$. 
The $M$ sets of background data $\{\mathbf{y}_j^k\}_{k=1}^M$  contain latent background subspace vectors that are also present  in $\{\mathbf{x}_i\}$.

Let $\mathbf{C}_{xx}:=m^{-1}\sum_{i=1}^{m}\mathbf{x}_i\mathbf{x}_i^\top$ and $\mathbf{C}_{yy}^k:=n_k^{-1}\times $ $
\sum_{j=1}^{n_k}\mathbf{y}_{j}^k(\mathbf{y}^k_{j})^\top$ be the corresponding sample covariance matrices. The goal here is to unveil  the latent subspace vectors 
\textcolor{black}{that are significant in representing the target data, but not any of the background data.}
Building on the dPCA in \eqref{eq:dpcam} for a single background dataset, it is meaningful to seek directions that maximize the variance of target data, while minimizing those of all background data. Formally, 
we pursue the following optimization, that we term multi-background (M) dPCA here, for discriminative analytics of multiple datasets
\begin{equation}\label{eq:gdpca}
\underset{\mathbf{U}\in\mathbb{R}^{D\times d}}{\max}	\quad {\rm Tr}\bigg[\bigg(\sum_{k=1}^M\omega_k\mathbf{U}^\top\mathbf{C}_{yy}^{k}\mathbf{U}\bigg)^{-1}\mathbf{U}^\top\mathbf{C}_{xx}\mathbf{U}\bigg]
\end{equation}
where $\{\omega_k\ge 0\}_{k=1}^M$ with $\sum_{k=1}^M \omega_k=1$
weight the variances of the $M$ projected background datasets.

Upon defining $\mathbf{C}_{yy}:=\sum_{k=1}^{M}\omega_k\mathbf{C}_{yy}^k
$, it is straightforward to see that \eqref{eq:gdpca} reduces to \eqref{eq:dpcam}. Therefore, one readily deduces 
 that the optimal ${\mathbf{U}}$ in \eqref{eq:gdpca} can be obtained by taking the $d$ right eigenvectors of $\mathbf{C}_{yy}^{-1}\mathbf{C}_{xx}$ that are associated with the $d$ largest eigenvalues. For implementation, the steps of MdPCA are presented in Alg. \ref{alg:mdpca}.

\begin{algorithm}[t]
	\caption{Multi-background dPCA.}
	\label{alg:mdpca}
	\begin{algorithmic}[1]
		\STATE {\bfseries Input:}
		Target  data $\{\accentset{\circ}{\mathbf{x}}_i\}_{i=1}^m$ and background data $\{\accentset{\circ}{\mathbf{y}}_{j}^k\}_{j=1}^{n_k}$ for $k=1,\,\ldots,\,M$; weight hyper-parameters $\{\omega_k\}_{k=1}^M$; number of dPCs $d$.
		\STATE {\bfseries Remove} the means from $\{\accentset{\circ}{\mathbf{x}}_i\}$ and $\{\accentset{\circ}{\mathbf{y}}_{j}^k\}_{k=1}^{M}$ to obtain  $\{\mathbf{x}_i\}$ and $\{\mathbf{y}_{j}^k\}_{k=1}^{M}$. Form $\mathbf{C}_{xx}$, $\{\mathbf{C}_{yy}^k\}_{k=1}^M$, and $\mathbf{C}_{yy}:=\sum_{k=1}^{M}\omega_k\mathbf{C}^k_{yy}$.
		\STATE {\bfseries Perform} eigendecomposition
		on $\mathbf{C}_{yy}^{-1}\mathbf{C}_{xx}$  to obtain the first $d$ right eigenvectors $\{\hat{\mathbf{u}}_i\}_{i=1}^d$.
		\STATE {\bfseries Output} $\hat{\mathbf{U}}:=[\hat{\mathbf{u}}_1\,\cdots\,\hat{\mathbf{u}}_d]$.
		\vspace{-0pt}
	\end{algorithmic}
\end{algorithm}

\begin{remark}
	\label{rmk:alpha}	
	The  parameters $\{\omega_k\}_{k=1}^M$ can be decided using two possible methods:
	 i) spectral-clustering \cite{ng2002spectral} to select a few sets of $\{\omega_k\}$ yielding \textcolor{black}{the most  representative subspaces for projecting the target data across  $\{\omega_k\}$;}
	 or ii) optimizing  $\{\omega_k\}_{k=1}^M$ jointly with $\mathbf{U}$ in \eqref{eq:gdpca}.
\end{remark}

For data  belonging to nonlinear manifolds, kernel (K) MdPCA will be developed next.
 With some nonlinear function  $\phi(\cdot)$, we obtain the transformed target data $\{\bm{\phi}(\mathbf{x}_i)\in\mathbb{R}^L\}$ as well as background data $\{\bm{\phi}(\mathbf{y}_{j}^k)\in\mathbb{R}^L\}$. Letting $\bm{\mu}_x\in\mathbb{R}^L$ and $\bm{\mu}_{y}^k:=(1/n_k)\sum_{j=1}^{n_k}\bm{\phi}(\mathbf{y}_{j}^k)\in\mathbb{R}^L$ denote the means of $\{\bm{\phi}(\mathbf{x}_i)\}$ and $\{\bm{\phi}(\mathbf{y}_{j}^k)\}$, respectively, one can form the corresponding covariance matrices $\mathbf{C}_{xx}^{\phi}\in\mathbb{R}^{L\times L}$, and 
\begin{equation*}
	\mathbf{C}_{yy}^{\phi,k}:=\frac{1}{n_k}\sum_{j=1}^{n_k}\left(\bm{\phi}(\mathbf{y}_{j}^k)-\bm{\mu}_{y}^k\right)\left(\bm{\phi}(\mathbf{y}_{j}^k)-\bm{\mu}_{y}^k\right)^\top\in\mathbb{R}^{L\times L}
	\end{equation*}
for $k=1,\,\ldots,\, M$.
Define the aggregate vector $\mathbf{b}_i\in\mathbb{R}^L$ 
\begin{equation*}
\label{eq:b}
\mathbf{b}_i:=\left\{
\begin{array}{ll}
\bm{\phi}(\mathbf{x}_i)-\bm{\mu}_x, & 1\le i \le m\\
\bm{\phi}(\mathbf{y}_{i-m}^1)-\bm{\mu}_{y}^1, & m< i \le m+n_1\\
~~~\vdots & \\
\bm{\phi}(\mathbf{y}_{i-\textcolor{black}{(N-n_M)}}^M)-\bm{\mu}_y^M,& N-n_{M}< i\le N
\end{array}
\right.
\end{equation*}
where $N:=m+\sum_{k=1}^M n_k$, for $i=1,\,\ldots,\,N$, and collect vectors $\{\mathbf{b}_i\}_{i=1}^N$ as columns to form $\mathbf{B}:=[\mathbf{b}_1\,\cdots\,\mathbf{b}_N]\in\mathbb{R}^{L\times N}$.
Upon assembling dual vectors $\{\mathbf{a}_i\in\mathbb{R}^N\}_{i=1}^d$ to form $\mathbf{A}:=[\mathbf{a}_1\,\cdots \, \mathbf{a}_d]\in\mathbb{R}^{N\times d}$,
the kernel version of \eqref{eq:gdpca} can be obtained as
	\begin{equation*}
	%\label{eq:kmdpcao}
\underset{\mathbf{A}\in\mathbb{R}^{N\times d}}{\max} {\rm Tr}\bigg[\bigg(\mathbf{A}^\top\mathbf{B}^\top\sum_{k=1}^M\omega_k\mathbf{C}^{\phi,k}_{yy}\mathbf{B}\mathbf{A}\hspace{-0.1cm}\bigg)^{-1}\mathbf{A}^\top\mathbf{B}^\top\mathbf{C}^{\phi} _{xx}\mathbf{B}\mathbf{A}\bigg].
	\end{equation*}

Consider now  kernel matrices  $\mathbf{K}_{xx}\in\mathbb{R}^{m\times m}$ and $\mathbf{K}_{kk}\in\mathbb{R}^{n_k\times n_k}$, whose $(i,\,j)$-th entries are   $\kappa(\mathbf{x}_i,\,\mathbf{x}_j)$ and $\kappa(\mathbf{y}_{i}^k,\,\mathbf{y}_{j}^k)$, respectively, for  $k=1,\,\ldots,\,M$. 
 Furthermore, matrices $\mathbf{K}_{xk}\in\mathbb{R}^{m\times n_k}$, and $\mathbf{K}_{lk}\in\mathbb{R}^{n_l\times n_k}$ are defined with  their corresponding $(i,\,j)$-th elements  $\kappa(\mathbf{x}_{i},\,\mathbf{y}_{j}^k)$ and $\kappa(\mathbf{y}_{i}^{l},\,\mathbf{y}_{j}^k)$, for $l=1,\,\ldots,\, k-1$ and  $k=1,\,\ldots,\,M$.
We subsequently center those matrices to obtain $\mathbf{K}_{xx}^c$ and 
\begin{align*}
\mathbf{K}_{kk}^c&:=\mathbf{K}_{kk}-\tfrac{1}{n_k}\mathbf{1}_{n_k}\mathbf{K}_{kk}-\tfrac{1}{n_k}\mathbf{K}_{kk}\mathbf{1}_{n_k}+\tfrac{1}{n_k^2}\mathbf{1}_{n_k}\mathbf{K}_{kk}\mathbf{1}_{n_k}\\
\mathbf{K}_{xk}^c&:=\mathbf{K}_{xk}-\tfrac{1}{m}\mathbf{1}_{m }\mathbf{K}_{xk}-\tfrac{1}{n_k}\mathbf{K}_{xk}\mathbf{1}_{n_k}+\tfrac{1}{mn_k}\mathbf{1}_{m}\mathbf{K}_{xk}\mathbf{1}_{n_k}\\
\mathbf{K}_{lk}^c&:=\mathbf{K}_{lk}-\!\tfrac{1}{n_l}\mathbf{1}_{n_l}\mathbf{K}_{lk}-\!\tfrac{1}{n_{k}}\mathbf{K}_{lk}\mathbf{1}_{n_k}+\!\tfrac{1}{n_ln_k}\mathbf{1}_{n_l}\mathbf{K}_{lk}\mathbf{1}_{n_k}
\end{align*}
where $\mathbf{1}_{n_k}\in\mathbb{R}^{n_k\times n_k}$ and $\mathbf{1}_{n_{l}}\in\mathbb{R}^{n_{l}\times n_{l}}$ are all-one matrices. With   $\mathbf{K}^x$ as in \eqref{eq:kx},  consider the $N\times N$ matrix
\begin{equation}
\label{eq:km}
\mathbf{K}:=\left[\begin{array}
{llll}
\mathbf{K}_{xx}^c  & \mathbf{K}_{x1}^c & \cdots & \mathbf{K}_{xM}^c \\
(\mathbf{K}_{x1}^c)^\top & \mathbf{K}_{11}^c  & \cdots & \mathbf{K}_{1M}^c \\
\quad\vdots           & \quad\vdots           & \ddots & \quad\vdots \\
(\mathbf{K}_{xM}^c)^\top & (\mathbf{K}_{1M}^c)^\top & \cdots & \mathbf{K}_{MM}^c
\end{array}
\right]
\end{equation}
and $\mathbf{K}^k\in\mathbb{R}^{N\times N}$ with  $(i,\,j)$-th entry
\begin{equation}
K^k_{i,j}:=\left\{\!\!\begin{array}{cl}
K_{i,j}/n_k, \!&\text{if}~ m+\sum_{\ell=1}^{n_{k-1}}n_{\ell}< i \le m+\sum_{\ell=1}^{n_{k}}n_{\ell}\\
0, &\mbox{otherwise}
\end{array}
\right.\label{eq:kk}
\end{equation}
for $k=1,\,\ldots,\,M$. Adopting the regularization in \eqref{eq:kdpcafm2}, our KMdPCA finds
	\begin{equation*}
%	\label{eq:kmdpca}
\hat{\mathbf{A}}:=\arg\underset{\mathbf{A}\in\mathbb{R}^{N\times d}}{\max}{\rm Tr}\bigg[\bigg(\mathbf{A}^\top\Big(\mathbf{K}\sum_{k=1}^M\mathbf{K}^k+\epsilon\mathbf{I}\Big)\mathbf{A}\bigg)^{-1}\!\!\mathbf{A}^\top\mathbf{K}\mathbf{K}^x\mathbf{A}\bigg]
	\end{equation*}
similar to (K)dPCA, whose solution comprises the right eigenvectors associated with the \textcolor{black}{first $d$} largest  eigenvalues in
	\begin{equation}
		\label{eq:kmdpcasol}
		\bigg(\mathbf{K}\sum_{k=1}^M\mathbf{K}^k+\epsilon\mathbf{I}\bigg)^{-1}\mathbf{K}\mathbf{K}^x\hat{\mathbf{a}}_i=\hat{\lambda}_i\hat{\mathbf{a}}_i.
	\end{equation}

For implementation, KMdPCA is presented in Alg. \ref{alg:mkdpca}.
\begin{algorithm}[t]
	\caption{Kernel multi-background dPCA.}
	\label{alg:mkdpca}
	\begin{algorithmic}[1]
		\STATE {\bfseries Input:}
		Target data $\{\mathbf{x}_i\}_{i=1}^m$ and background data $\{\mathbf{y}_{j}^k\}_{j=1}^{n_k}$ for $k=1,\,\ldots,\,M$; number of dPCs $d$; kernel function $\kappa(\cdot)$; weight coefficients $\{\omega_k\}_{k=1}^M$; constant $\epsilon$.
		\STATE {\bfseries Construct} $\mathbf{K}$ using \eqref{eq:km}. Build $\mathbf{K}^x$ and $\{\mathbf{K}^k\}_{k=1}^M$ via \eqref{eq:kx} and \eqref{eq:kk}.
		\STATE {\bfseries Solve} \eqref{eq:kmdpcasol} to obtain the first $d$  eigenvectors $\{\hat{\mathbf{a}}_i\}_{i=1}^d$.
		\STATE {\bfseries Output} $\hat{\mathbf{A}}:=[\hat{\mathbf{a}}_1\,\cdots\,\hat{\mathbf{a}}_d]$. 
		\vspace{-0pt}
	\end{algorithmic}
\end{algorithm}

\begin{remark}
We can verify that PCA, KPCA, dPCA, KdPCA, MdPCA, and KMdPCA incur computational complexities $\mathcal{O}(mD^2)$, $\mathcal{O}(m^2D)$, $\mathcal{O}(\max(m,n)D^2)$, $\mathcal{O}(\max(m^2,n^2)D)$, $\mathcal{O}(\max(m,\bar{n} )D^2)$, and $\mathcal{O}(\max(m^2,\bar{n}^2 )D)$, respectively, where $\bar{n}:=\max_k~\{n_k\}_{k=1}^M$. 
It is also not difficult to check that the computational complexity of 
	forming $\mathbf{C}_{xx}$, $\mathbf{C}_{yy}$,  $\mathbf{C}_{yy}^{-1}$, and performing the eigendecomposition on $\mathbf{C}_{yy}^{-1}\mathbf{C}_{xx}$ is $\mathcal{O}(mD^2)$, $\mathcal{O}(nD^2)$, $\mathcal{O}(D^3)$, and $\mathcal{O}(D^3)$, respectively. As the number of data vectors ($m,\,n$) is much larger than their dimensionality $D$, when performing dPCA in the primal domain, it follows readily that dPCA incurs  complexity $\mathcal{O}(\max(m,n)D^2)$. Similarly, the computational complexities of the other algorithms can be checked.% verified.
%}
Evidently,  when $\min(m,n)\gg D $ or $\min(m,\underline{n})\gg D$ with $\underline{n}:=\min_k\,\{n_k\}_{k=1}^M$,
	dPCA and MdPCA are computationally more attractive than KdPCA and KMdPCA. On the other hand, KdPCA and KMdPCA become more appealing, when $D\gg \max(m,n)$ or $D\gg \max(m,\bar{n})$.
	Moreover, the computational complexity of cPCA is $\mathcal{O}(\max (m,n)D^2L)$, where $L$ denotes the number of $\alpha$'s candidates. Clearly, relative to dPCA, cPCA is computationally more expensive when $DL> \max(m,n)$.
\end{remark}

\section{Numerical Tests}\label{sec:simul}

To evaluate the performance of our proposed approaches for discriminative analytics, we carried out a number of  numerical tests using several synthetic and real-world datasets, a sample of which are reported in this section. 

\subsection{dPCA tests}

\begin{figure}[t]
 	\centering 
 	\includegraphics[scale=0.64]{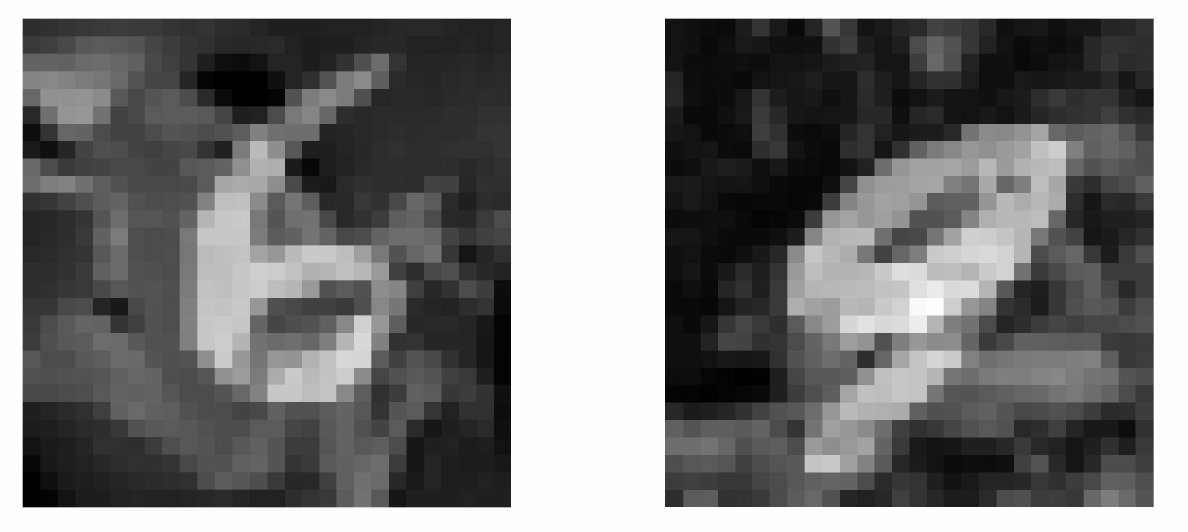} 
 	\caption{\small{Superimposed images.}}
 	\label{fig:targ}
 \end{figure}

 Semi-synthetic target \textcolor{black}{$\{\accentset{\circ}{\mathbf{x}}_i\in\mathbb{R}^{784}\}_{i=1}^{2,000}$} and background images \textcolor{black}{$\{\accentset{\circ}{\mathbf{y}}_j\in\mathbb{R}^{784}\}_{j=1}^{3,000}$} were obtained by superimposing images from the MNIST \footnote{Downloaded from http://yann.lecun.com/exdb/mnist/.} and CIFAR-10 \cite{cifar10} datasets.
 Specifically, the target data $\{\mathbf{x}_i\in\mathbb{R}^{784}\}_{i=1}^{2,000}$ were generated using $2,000$ handwritten digits 6 and 9 (1,000 for each) of size $28\times 28$,
  superimposed with $2,000$ frog images from the CIFAR-10 database \cite{cifar10} followed by removing the sample mean from each data point; see Fig. \ref{fig:targ}. The raw $32\times 32$ frog images were converted into grayscale, and randomly cropped to $28\times 28$. The zero-mean background data $\{\mathbf{y}_j\in\mathbb{R}^{784}\}_{j=1}^{3,000}$ were constructed using $3,000$ cropped frog images, which were randomly chosen from the remaining frog images in the CIFAR-10 database.

\begin{figure}[t]
	\centering 
	\includegraphics[scale=0.53]{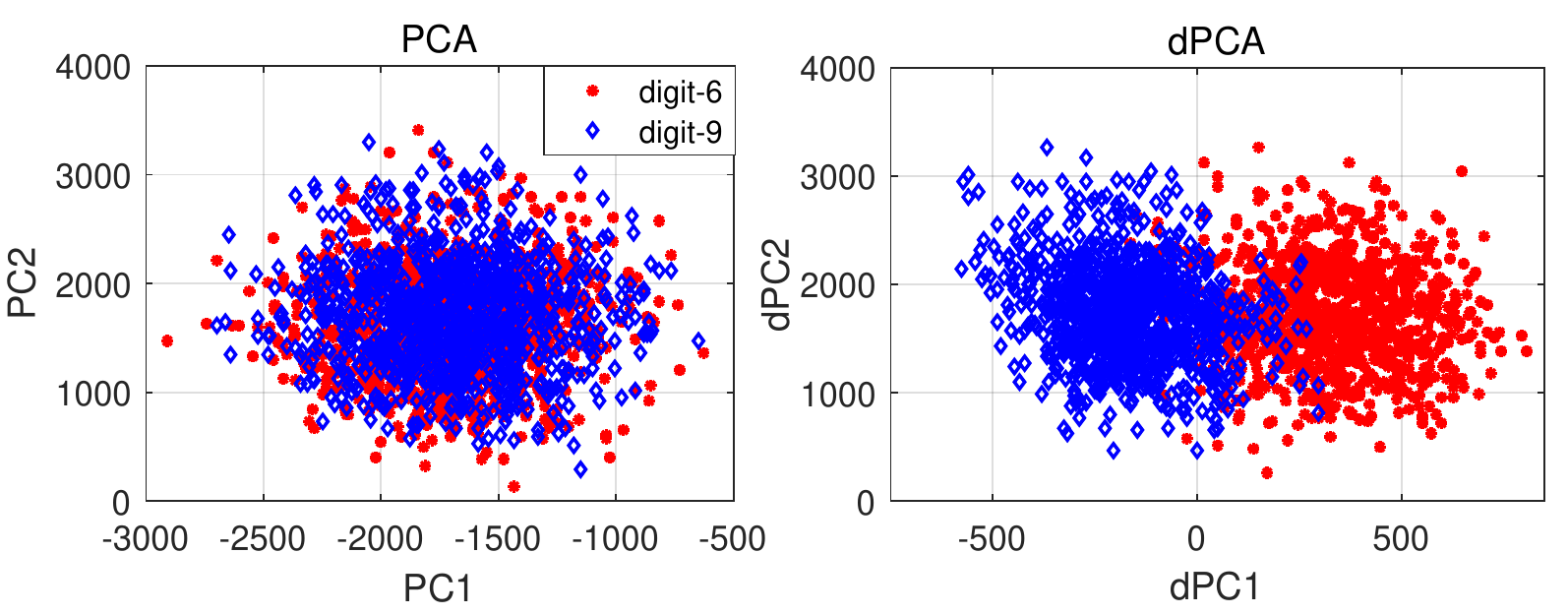} 	
	\caption{\small{dPCA versus PCA on semi-synthetic images.}}
	\label{fig:digits}
\end{figure}

The dPCA Alg. \ref{alg:dpca} was performed on $\{\accentset{\circ}{\mathbf{x}}_i\}$ and  $\{\accentset{\circ}{\mathbf{y}}_j\}$ with $d=2$. PCA was implemented on $\{\accentset{\circ}{\mathbf{x}}_i\}$ only. The first two PCs and dPCs are presented in the left and right panels of Fig. \ref{fig:digits}, respectively. Clearly, dPCA reveals the discriminative information of the target data describing digits $6$ and $9$ relative to the background data, enabling successful discovery of the digit $6$ and $9$ subgroups. On the contrary, PCA captures only the patterns that correspond to the generic background rather than those associated with  the digits $6$ and $9$. 
\textcolor{black}{To further assess the performance of dPCA and PCA, K-means is carried out using the resulting low-dimensional representations of the target data. The clustering performance is evaluated in terms of two metrics: clustering error and scatter ratio. The clustering error is defined as the ratio of the number of incorrectly clustered data vectors over $m$.
	Scatter ratio verifying cluster separation is defined as $S_t/\sum_{i=1}^2S_i$, where $S_t$ and $\{S_i\}_{i=1}^2$ denote the total scatter value and the within cluster scatter values, given by  $S_t:=\sum_{j=1}^{2,000}\|\hat{\mathbf{U}}^\top\mathbf{x}_j\|_2^2$ and $\{S_i:=\sum_{j\in \mathcal{C}_i}\|\hat{\mathbf{U}}^\top\mathbf{x}_j-\hat{\mathbf{U}}^\top\sum_{k\in{\mathcal{C}}_i}\mathbf{x}_k\|_2^2\}_{i=1}^2$, respectively, with $\mathcal{C}_i$ representing the set of data vectors belonging to cluster $i$. Table \ref{tab:cluster} reports the clustering errors and scatter ratios of dPCA and PCA under  different $d$ values. Clearly, dPCA exhibits lower clustering error and higher scatter ratio.}

\renewcommand{\arraystretch}{1.5} 
\begin{table}[tp]	
\centering
	\fontsize{8.5}{8}\selectfont
		\textcolor{black}{\caption{\textcolor{black}{Performance comparison between dPCA and PCA.}}
	\label{tab:cluster}
	\vspace{.8em}
	\begin{tabular}{|c|c|c|c|c|}
		\hline
		\multirow{2}{*}{$d$}&
		\multicolumn{2}{c|}{Clustering error}&\multicolumn{2}{c|}{ Scatter ratio}\cr\cline{2-5}
		&dPCA&PCA&dPCA&PCA\cr
		\hline
		\hline
		1&0.1660&0.4900&2.0368&1.0247\cr\hline
		2&0.1650&0.4905&1.8233&1.0209\cr\hline
		3&0.1660&0.4895&1.6719&1.1327\cr\hline
		4&0.1685&0.4885&1.4557&1.1190\cr\hline
		5&0.1660&0.4890&1.4182&1.1085\cr\hline
		10&0.1680&0.4885&1.2696&1.0865\cr\hline
		50&0.1700&0.4880&1.0730&1.0568\cr\hline
	   100&0.1655&0.4905&1.0411&1.0508\cr
		\hline
	\end{tabular}}
\end{table}

Real protein expression data \cite{mice}
 were also used to evaluate the ability of dPCA to discover subgroups in real-world conditions. Target data $\{\accentset{\circ}{\mathbf{x}}_i\in\mathbb{R}^{77}\}_{i=1}^{267}$ contained $267$ data vectors, each collecting $77$ protein expression measurements of a mouse having Down Syndrome disease \cite{mice}.  
In particular, the first $135$ data points $\{\accentset{\circ}{\mathbf{x}}_i\}_{i=1}^{135}$ recorded protein expression measurements of $135$ mice with drug-memantine treatment, while the remaining $\{\accentset{\circ}{\mathbf{x}}_i\}_{i=136}^{267}$ collected measurements of $134$ mice without such treatment. Background data  $\{\accentset{\circ}{\mathbf{y}}_j\in\mathbb{R}^{77}\}_{j=1}^{135}$ on the other hand, comprised such measurements from $135$ healthy mice, which likely exhibited similar natural variations (due to e.g., age and sex) as the target mice, but without the differences that result from the Down Syndrome disease.

When performing cPCA on $\{\accentset{\circ}{\mathbf{x}}_i\}$ and $\{\accentset{\circ}{\mathbf{y}}_j\}$, four $\alpha$'s were selected from $15$ logarithmically-spaced values between $10^{-3}$ and $10^{3}$ via the spectral clustering method presented in \cite{2017cpca}.

\begin{figure}[t]
	\centering 
	\includegraphics[scale=0.735]{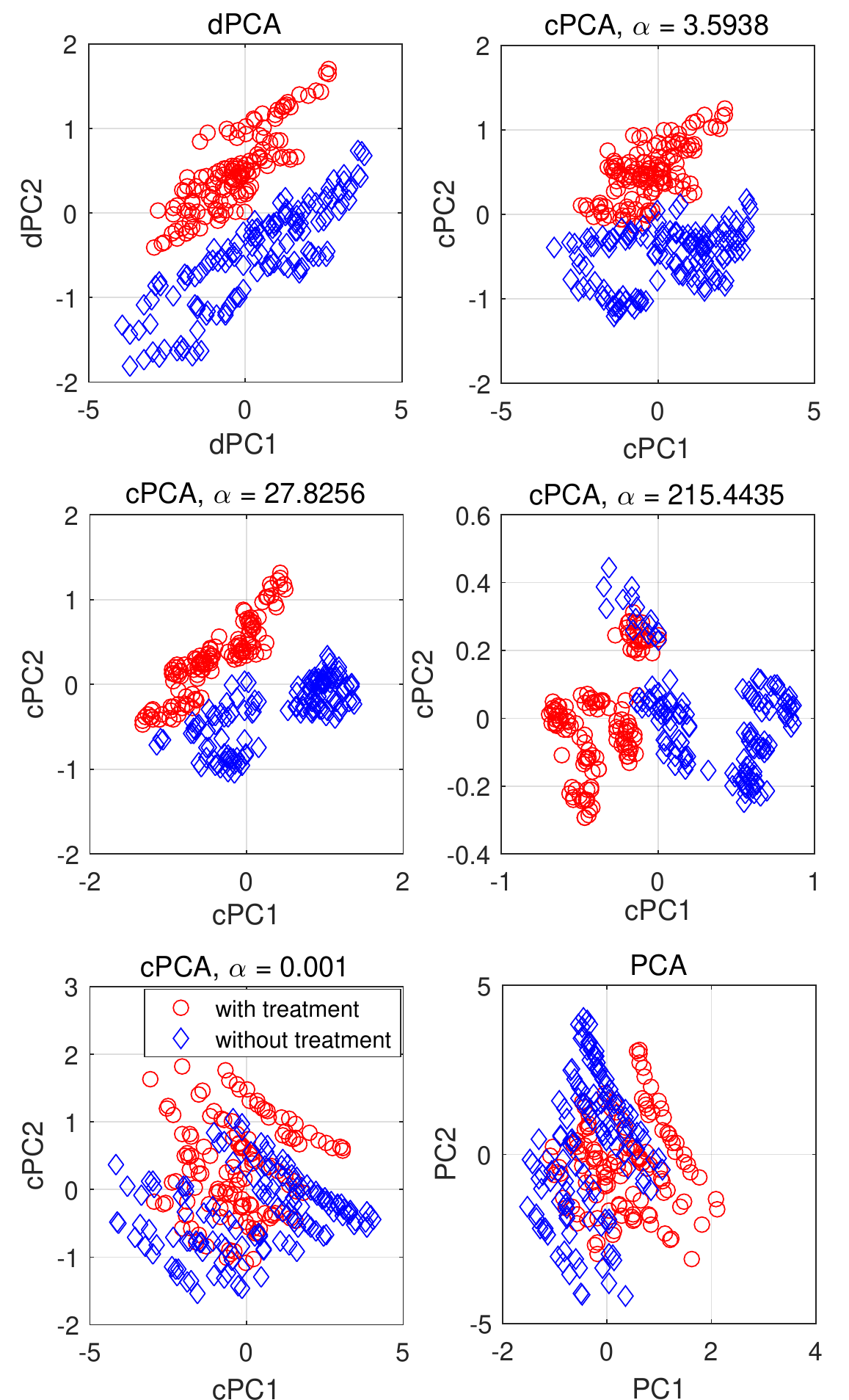} 
	\vspace{-4pt}
	\caption{\small{Discovering subgroups in mice protein expression data.}}
	\label{fig:mice}
	\vspace{-4pt}
\end{figure}

Experimental results are reported in Fig. \ref{fig:mice} with red circles and black diamonds representing sick mice with and without treatment, respectively. Evidently, when PCA is applied, the low-dimensional representations of the protein measurements from mice with and without treatment are distributed similarly. In contrast,  the low-dimensional representations cluster two groups of mice successfully when dPCA is employed. At the price of runtime (about $15$ times more than dPCA), cPCA with well {tuned} parameters ($\alpha=3.5938$ and $27.8256$) can also separate the two groups.

\subsection{KdPCA tests}\label{sec:simu2}

 \begin{figure}[t]
 	\centering 
 	\includegraphics[scale=0.61]{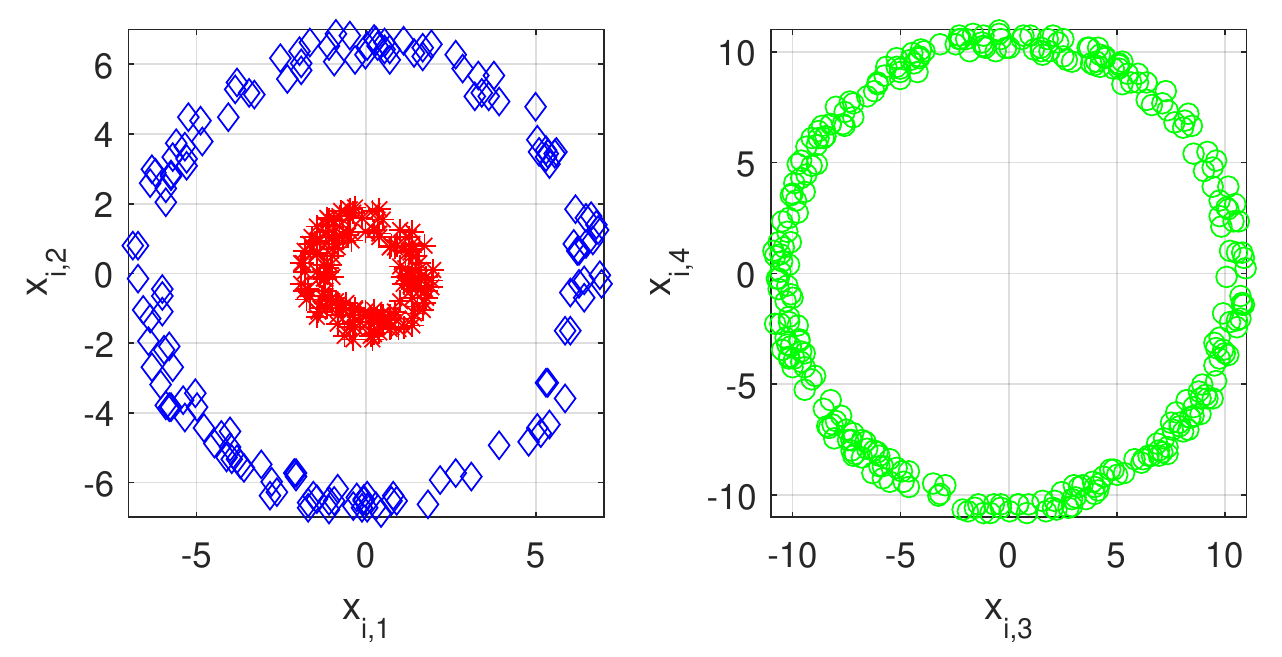} 
 	\caption{\small{Target data dimension distributions with $x_{i,j}$ representing the $j$-th entry of   $\mathbf{x}_i$ for $j=1,\, \ldots,\,4$ and $i=1,\, \ldots,\,300$.}}
 	\label{fig:kdpca_target}
 \end{figure}
 
 In this subsection, our KdPCA is evaluated 
% compared against the state-of-the-art 
 using synthetic and real data.
By adopting the procedure described in \cite[p. 546]{hastie2009elements}, we generated target data $\{{\mathbf{x}}_i:=[x_{i,1}\, x_{i,2}\,x_{i,3}\,x_{i,4}]^\top\}_{i=1}^{300}$ and background data $\{{\mathbf{y}}_j\in\mathbb{R}^4\}_{j=1}^{150}$. In detail, $\{[x_{i,1}\,x_{i,2}]^{\top}\}_{i=1}^{300}$ were sampled uniformly  from two circular concentric clusters with corresponding radii $1$ and $6$ shown in the left panel of Fig. \ref{fig:kdpca_target}; and $\{[x_{i,3}\,x_{i,4}]^{\top}\}_{i=1}^{300}$ were uniformly drawn from a circle with radius $10$; see Fig. \ref{fig:kdpca_target} (right panel) for illustration.
 The first and second two dimensions of $\{{\mathbf{y}}_j\}_{j=1}^{150}$ were uniformly sampled from two concentric circles with corresponding radii of $4$ and $10$. All data points in $\{{\mathbf{x}}_i\}$ and $\{{\mathbf{y}}_j\}$ were corrupted with additive noise sampled independently from $\mathcal{N}(\mathbf{0},\,0.1\mathbf{I})$. 
 To unveil the specific cluster structure of the target data relative to the background data,  
  Alg. \ref{alg:kdpca} was run with $\epsilon=10^{-3}$ and using the degree-$2$ polynomial kernel $\kappa(\mathbf{z}_i,\mathbf{z}_j)=(\mathbf{z}_i^\top\mathbf{z}_j )^2$.   Competing alternatives including PCA, KPCA, cPCA, kernel (K) cPCA \cite{2017cpca}, and dPCA were also implemented.  
  Further, KPCA and KcPCA shared the kernel function with KdPCA. Three different values of $\alpha$ were automatically chosen for cPCA \cite{2017cpca}.
  The parameter $\alpha$ of KcPCA was set as $1$, $10$, and $100$.

\begin{figure}[t]
	\centering 
	\includegraphics[scale=0.78]{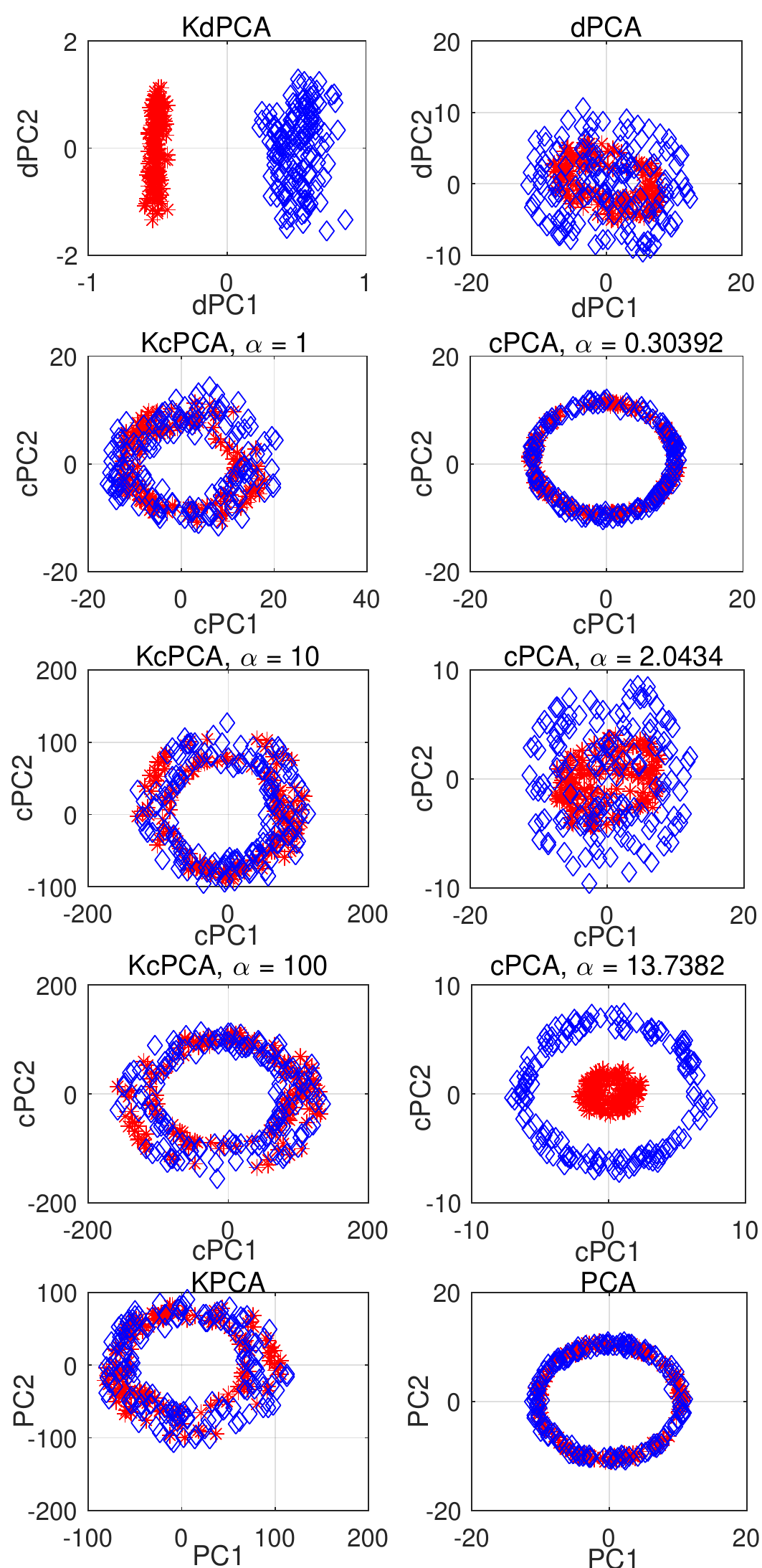}  
			\vspace{5pt}
	\caption{\small{Discovering subgroups in nonlinear synthetic data.}}
	\label{fig:kdpca_syn}
\end{figure}

Figure \ref{fig:kdpca_syn} depicts the first two dPCs, cPCs, and PCs of the aforementioned dimensionality reduction algorithms. Clearly, only KdPCA
successfully reveals the two unique clusters of $\{{\mathbf{x}}_i\}$ relative to $\{{\mathbf{y}}_j\}$. 

\begin{figure}[t]
	\centering 
	\includegraphics[scale=0.74]{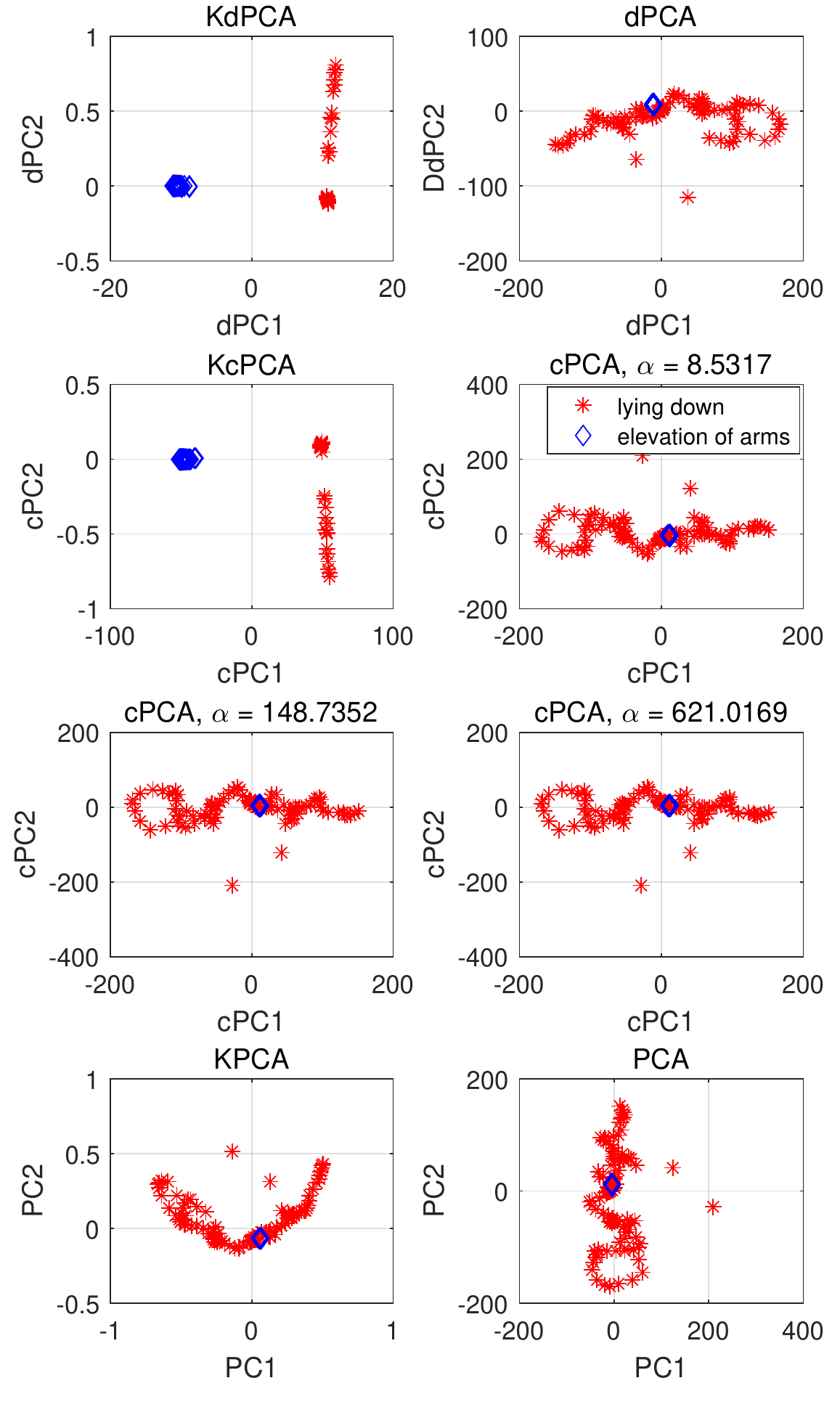}  
	\caption{\small{Discovering subgroups in MHealth data.}}
	\label{fig:kdpca_real}
\end{figure}

\begin{figure}[t]
	\centering 
	\includegraphics[scale=0.71]{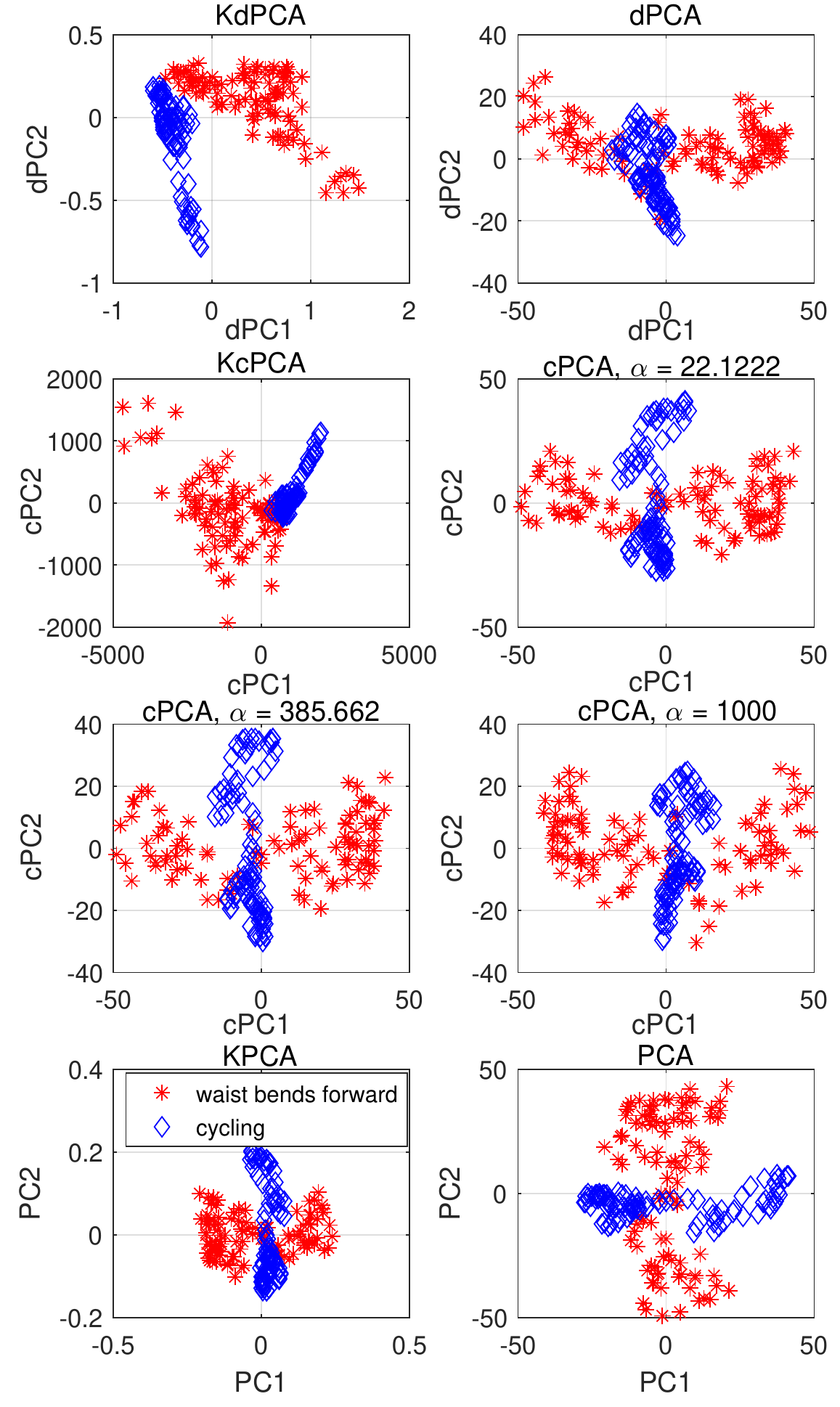}  
	\caption{\small{Distinguishing between waist bends forward and cycling.}}
	\label{fig:kdpca_real_test2} 
\end{figure}

\begin{figure}[t]
	\centering 
	\includegraphics[scale=0.7]{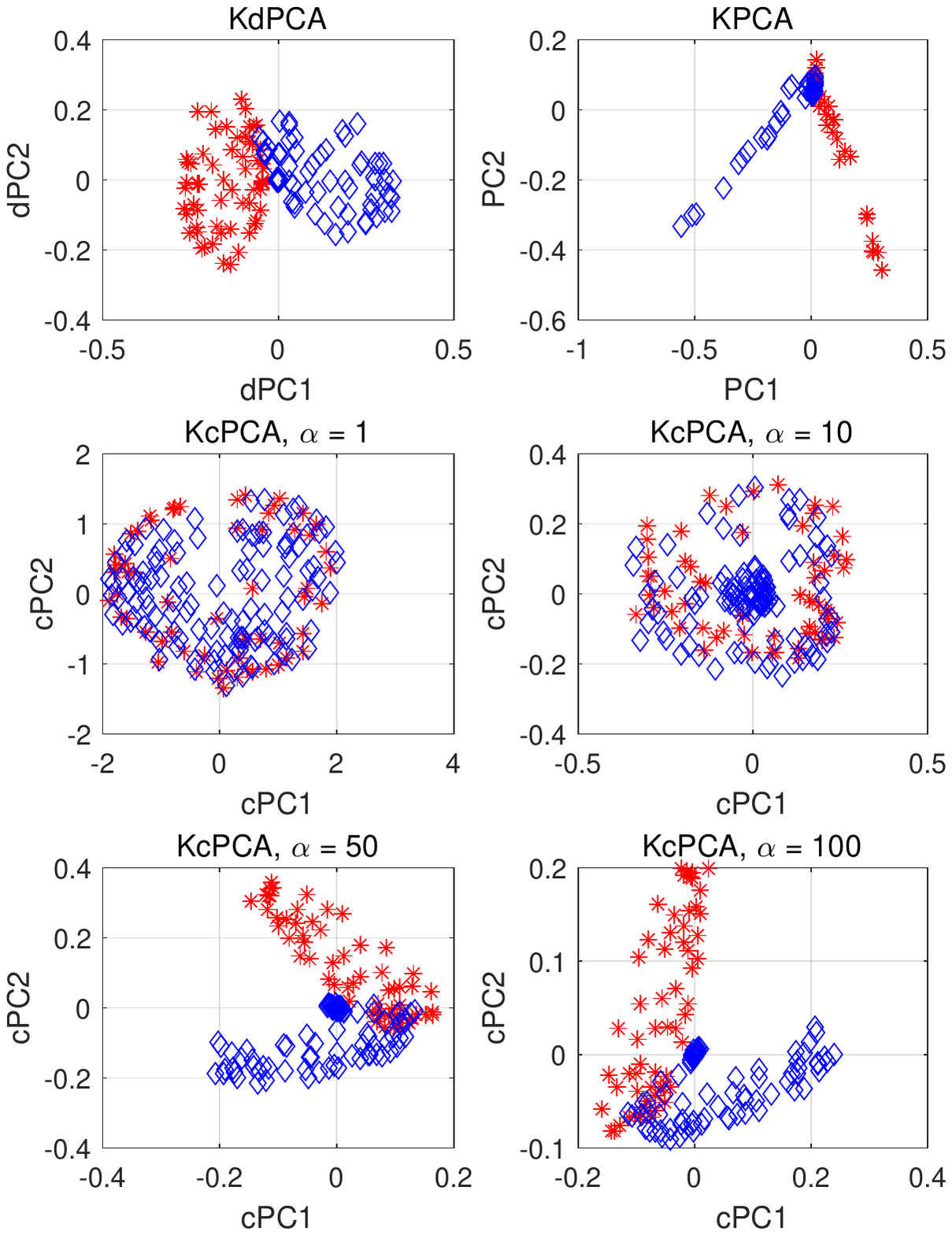}  
	\caption{\small{Face recognization by performing KdPCA.}}
	\label{fig:kdpca_real_add} 
\end{figure}

KdPCA was tested 
in realistic settings using the real Mobile (M) Health data \cite{mhealth}. This dataset consists of sensor (e.g., gyroscopes, accelerometers, and EKG) measurements from volunteers conducting a series of physical activities. In the first experiment, $200$ target data  $\{{\mathbf{x}}_i\in\mathbb{R}^{23}\}_{i=1}^{200}$ were used, each of which recorded $23$ sensor measurements from one volunteer performing two different physical activities, namely laying down and having  frontal elevation of arms ($100$ data points correspond to each activity). Sensor measurements from the same volunteer standing still were utilized for the $100$ background data points $\{{\mathbf{y}}_j\in\mathbb{R}^{23}\}_{j=1}^{100}$. 
For KdPCA, KPCA, and KcPCA algorithms, the Gaussian kernel  
 with bandwidth $5$
was used. Three different values for the parameter $\alpha$ in cPCA were automatically selected from a list of $40$ logarithmically-spaced values between $10^{-3}$ and $10^{3}$, whereas $\alpha$ in KcPCA was set to $1$ \cite{2017cpca}.

The first two dPCs, cPCs, and PCs of KdPCA, dPCA, KcPCA, cPCA, KPCA, and PCA are reported in Fig. \ref{fig:kdpca_real}. It is self-evident that the two activities evolve into two separate clusters in the plots of KdPCA and KcPCA. On the contrary, due to the nonlinear data correlations, the other alternatives fail to distinguish the two activities.

In the second experiment, the target data were formed with sensor measurements of one volunteer executing waist bends forward and cycling. The background data were collected from the same volunteer standing still. The Gaussian kernel with bandwidth $40$ was used for KdPCA and KPCA, while the second-order polynomial kernel $\kappa(\mathbf{z}_i,\,\mathbf{z}_j)=(\mathbf{z}_i^\top\mathbf{z}_j+3)^2$ was employed for KcPCA.  The first two dPCs, cPCs, and PCs of simulated schemes are depicted
 in Fig. \ref{fig:kdpca_real_test2}. 
Evidently, KdPCA outperforms its competing alternatives in discovering the two physical activities of the target data.

\textcolor{black}{To test the scalability of our developed schemes,  the Extended Yale-B (EYB)  face image dataset \cite{yaleb} was adopted to test 
	 the clustering performance of KdPCA, KcPCA, and KPCA. EYB  database contains frontal face images of $38$ individuals, each having about around $65$ color images of $192\times 168$ ($32,256$) pixels. The color images of three individuals ($60$ images per individual) were converted into grayscale images and vectorized to obtain $180$ vectors of   size $32,256\times 1$. The $120$ vectors from two individuals (clusters) comprised the target data, and the remaining $60$ vectors formed the background data. A Gaussian kernel with bandwidth $150$ was used for KdPCA, KcPCA, and KPCA. Figure \ref{fig:kdpca_real_add} reports the first two dPCs, cPCs, and PCs of KdPCA, KcPCA (with 4 different values of $\alpha$), and KPCA, with black circles and red stars representing the two different individuals from the target data. K-means is carried out using the resulting $2$-dimensional representations of the target data. The clustering errors of KdPCA, KcPCA with $\lambda=1$, KcPCA with $\lambda=10$, KcPCA with $\lambda=50$, KcPCA with $\lambda=100$, and KPCA are $0.1417$, $0.7$, $0.525$, $0.275$, $0.2833$, and  $0.4167$, respectively. Evidently, the face images of the two individuals can be better recognized with KdPCA than with other methods.}

\subsection{MdPCA tests}

The ability of the MdPCA Alg. \ref{alg:mdpca} for discriminative dimensionality reduction  
is examined here with two background datasets. 
 For simplicity, the involved weights were set to $\omega_1=\omega_2=0.5$.	
 
\begin{figure}[t]
	\centering 
	\includegraphics[scale=0.66]{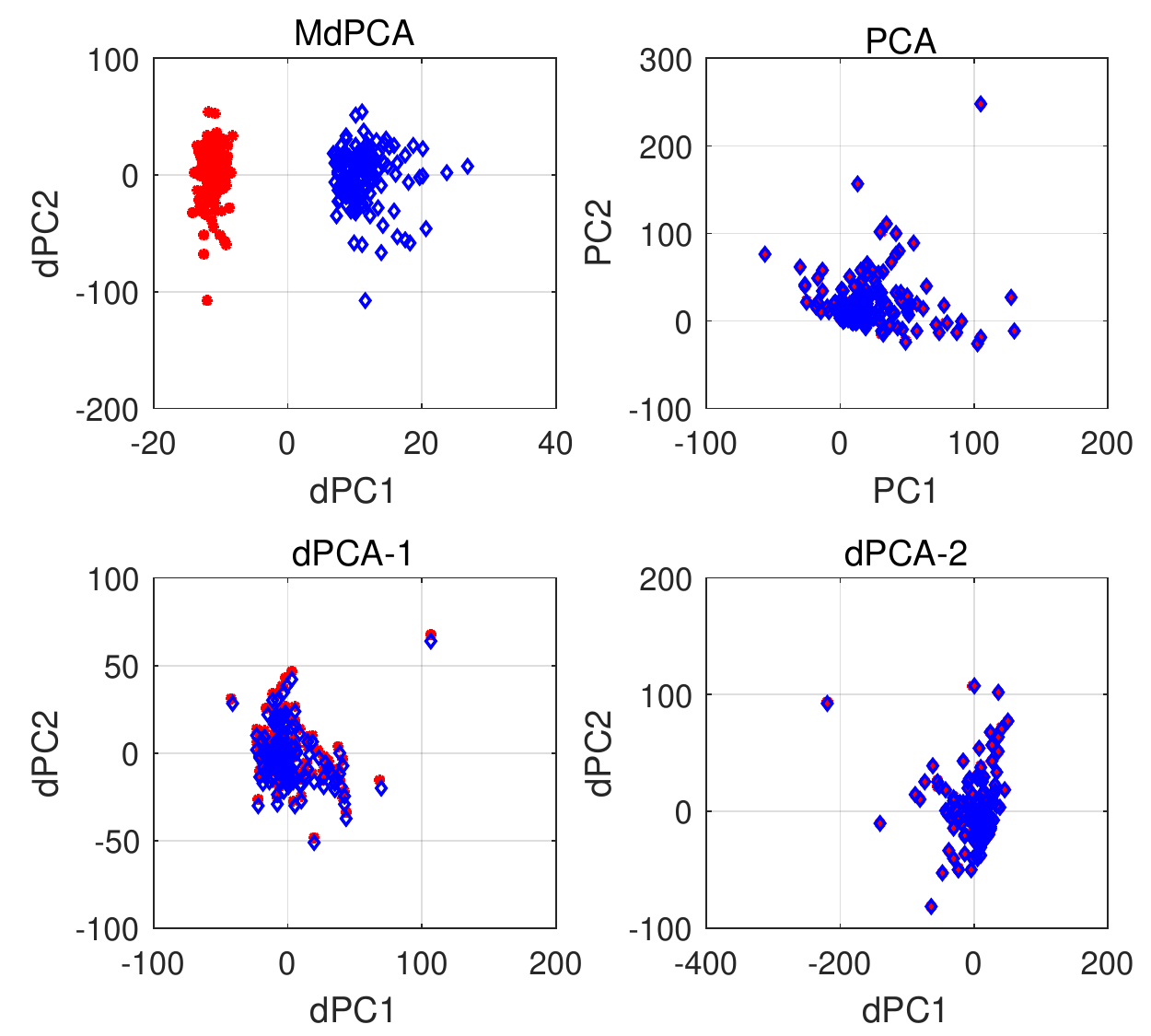}  
	\vspace{-5pt}
	\caption{\small{Clustering structure by MdPCA using synthetic data.}}
	\label{fig:mdpca_syn}
	\vspace{-5pt}
\end{figure}

 In the first experiment,
	two clusters of $15$-dimensional data points were generated for the target data $\{\accentset{\circ}{\mathbf{x}}_i\in\mathbb{R}^{15}\}_{i=1}^{300}$ ($150$ for each). 
Specifically, the first $5$ dimensions of $\{\accentset{\circ}{\mathbf{x}}_i\}_{i=1}^{150}$ and $\{\accentset{\circ}{\mathbf{x}}_i\}_{i=151}^{300}$ were sampled from $\mathcal{N}(\mathbf{0},\,\mathbf{I})$ and $\mathcal{N}(8\mathbf{1},\,2\mathbf{I})$, respectively. The second and last $5$ dimensions of $\{\accentset{\circ}{\mathbf{x}}_i\}_{i=1}^{300}$ were drawn accordingly from the normal distributions $\mathcal{N}(\mathbf{1},\,10\mathbf{I})$ and $\mathcal{N}(\mathbf{1},\,20\mathbf{I})$. The right top plot of Fig. \ref{fig:mdpca_syn} shows that performing PCA cannot resolve the two clusters.
The first, second, and last $5$ dimensions of the first background dataset $\{\accentset{\circ}{\mathbf{y}}_{j}^1\in\mathbb{R}^{1}\}_{j=15}^{150}$  were sampled from $\mathcal{N}(\mathbf{1},\,2\mathbf{I})$, $\mathcal{N}(\mathbf{1},\,10\mathbf{I})$, and $ \mathcal{N}(\mathbf{1},\,2\mathbf{I})$, respectively, while those of the second background dataset $\{\accentset{\circ}{\mathbf{y}}_{j}^2\in\mathbb{R}^{15}\}_{j=1}^{150}$ were drawn from $\mathcal{N}(\mathbf{1},\,2\mathbf{I})$, $\mathcal{N}(\mathbf{1},\,2\mathbf{I})$, and $\mathcal{N}(\mathbf{1},\,20\mathbf{I})$.  
The two plots at the bottom of Fig. \ref{fig:mdpca_syn} depict the first two dPCs of dPCA implemented with a single background dataset.  
Evidently, MdPCA can discover the two clusters in the target data by leveraging 
 the two background datasets.  

\begin{figure}[t]
	\centering 
	\includegraphics[scale=0.69]{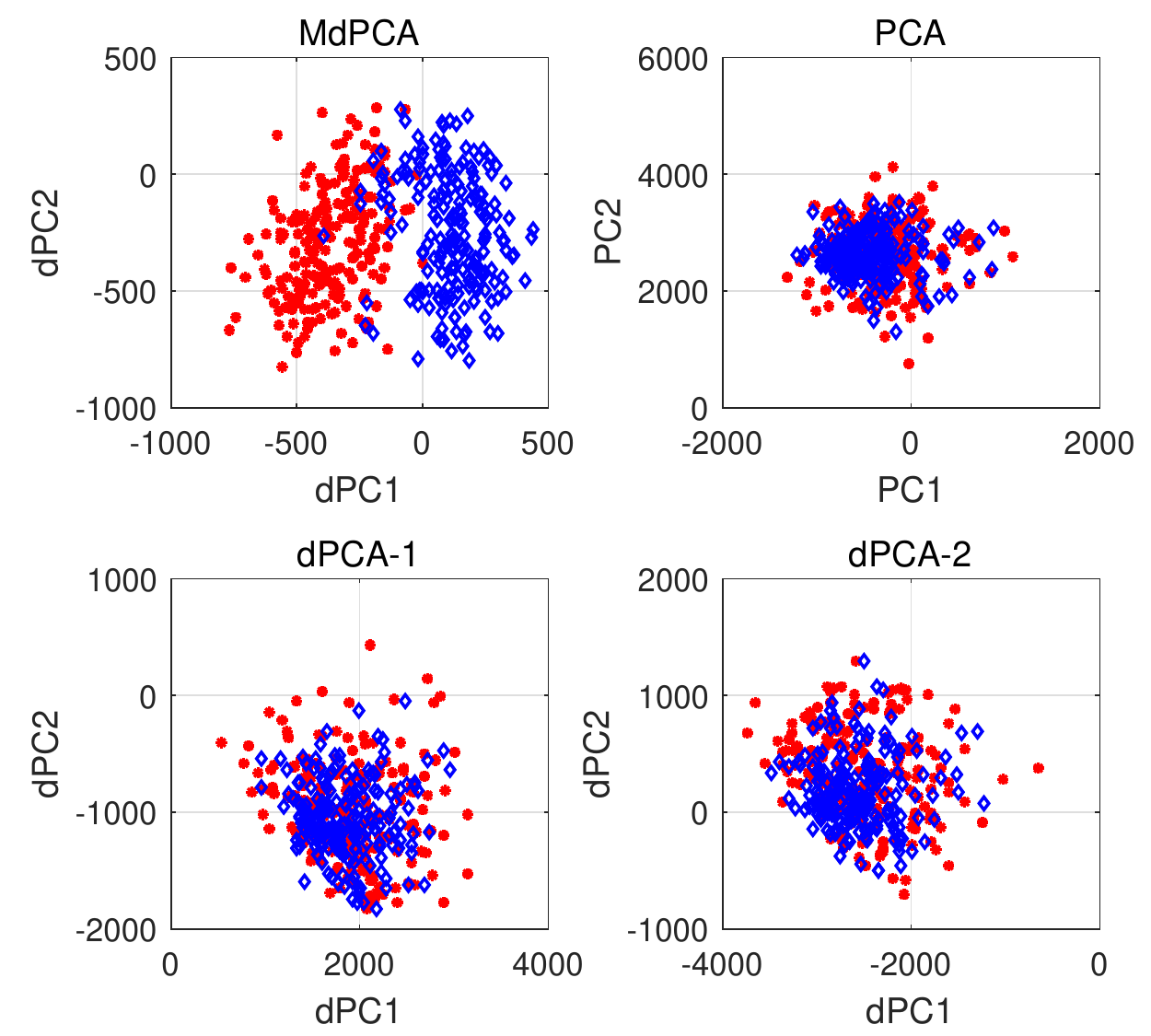}  
	\vspace{-8pt}
	\caption{\small{Clustering structure by MdPCA using semi-synthetic data.}}
	\label{fig:mdpca_semisyn}
	\vspace{-10pt}
\end{figure}

In the second experiment, the
  target data $\{\accentset{\circ}{\mathbf{x}}_{i}\in\mathbb{R}^{784}\}_{i=1}^{400}$ were obtained using $400$ handwritten digits $6$ and $9$ ($200$ for each) of size $28\times 28$ from the MNIST dataset superimposed with $400$ resized `girl' images from the CIFAR-100 dataset  \cite{cifar10}. 
The first $392$ dimensions of the first background dataset $\{\accentset{\circ}{\mathbf{y}}_{j}^1\in\mathbb{R}^{784}\}_{j=1}^{200}$ and the last $392$ dimensions of the other background dataset $\{\accentset{\circ}{\mathbf{y}}_{j}^2\in\mathbb{R}^{784}\}_{j=1}^{200}$ correspond to the first and last $392$ features of $200$ cropped girl images, respectively. The remaining dimensions of both background datasets were set zero.
 Figure \ref{fig:mdpca_semisyn} presents the obtained (d)PCs of MdPCA, dPCA, and PCA, with red stars and black diamonds depicting digits $6$ and $9$, respectively.  
 PCA and dPCA based on a single background dataset (the bottom two plots in Fig. \ref{fig:mdpca_semisyn}) reveal that the two clusters of data follow a similar distribution in the space spanned by the first two PCs. The separation between the two clusters becomes clear when the MdPCA is employed.

\subsection{KMdPCA tests}
\begin{figure}[t]
	\centering 
	\includegraphics[scale=0.7]{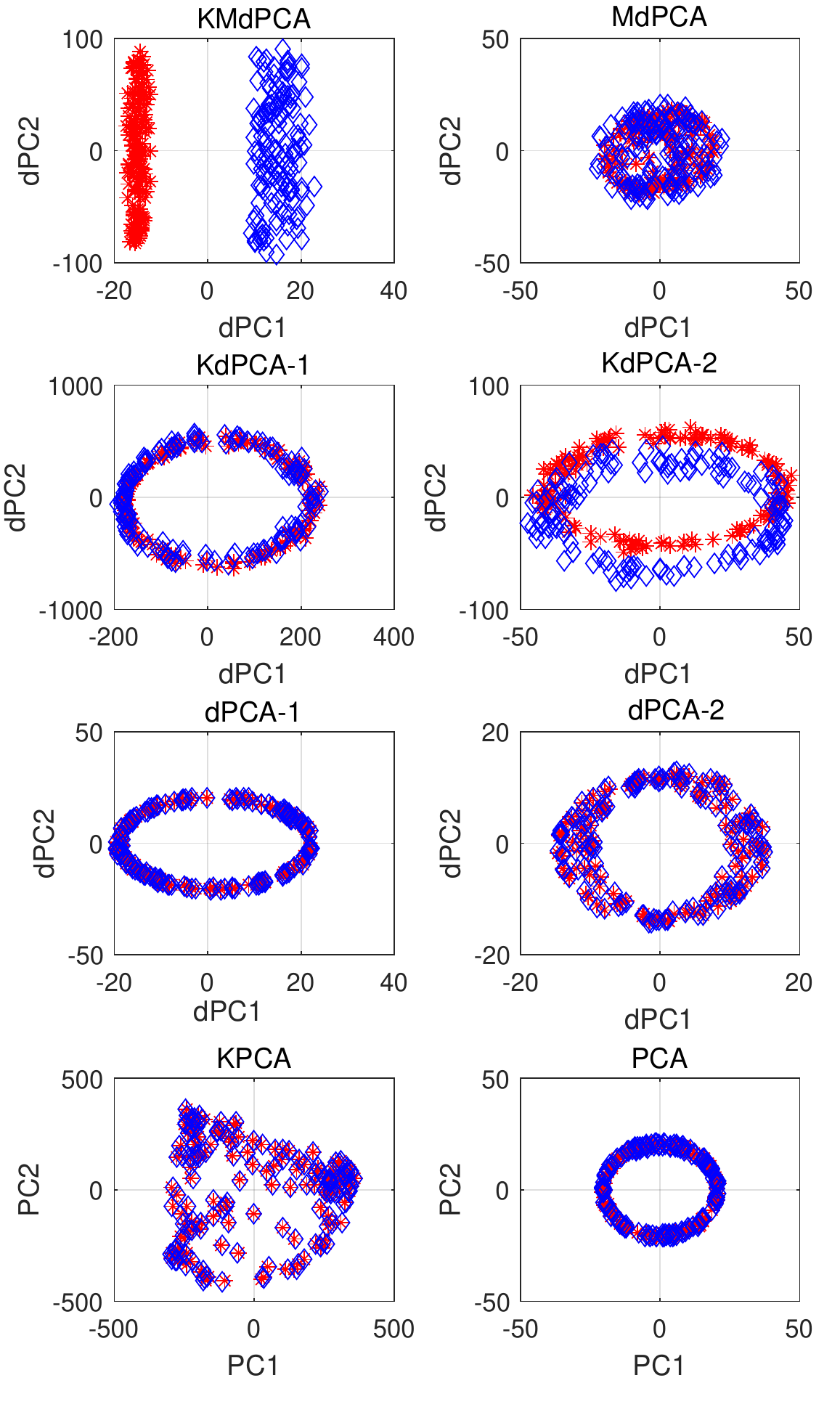} 
	\vspace{-10pt}
	\caption{\small{The first two dPCs obtained by  Alg. \ref{alg:mkdpca}.}}
	\label{fig:kmdpca}
	\vspace{-13pt}
\end{figure}

Algorithm \ref{alg:mkdpca} \textcolor{black}{with $\epsilon=10^{-4}$} is examined for dimensionality reduction using simulated data and compared against MdPCA, KdPCA, dPCA, and PCA.
The first two dimensions of the target data $\{\mathbf{x}_i\in \mathbb{R}^6\}_{i=1}^{150}$ and $\{\mathbf{x}_i\}_{i=151}^{300}$ were generated from two circular concentric clusters with respective radii  of $1$ and $6$.  The remaining four dimensions of the target data $\{\mathbf{x}_i\}_{i=1}^{300}$ were sampled from   two concentric circles with radii of $20$ and $12$, respectively. Data $\{\mathbf{x}_i\}_{i=1}^{150}$ and $\{\mathbf{x}_i\}_{i=151}^{300}$ corresponded to two different clusters. 
The first, second, and last two dimensions of one background dataset $\{\mathbf{y}_{j}^1\in\mathbb{R}^6\}_{j=1}^{150}$ were sampled from three concentric circles with corresponding radii of $3$, $3$, and $12$. 
Similarly,  three concentric circles with radii  $3$, $20$, and $3$ were used for generating the other background dataset $\{\mathbf{y}_{j}^2\in\mathbb{R}^6\}_{j=1}^{150}$. Each datum in $\{\mathbf{x}_i\}$, $\{\mathbf{y}_{j}^1\}$, and $\{\mathbf{y}_{j}^2\}$ was corrupted by additive noise $ \mathcal{N}(\mathbf{0}, \,0.1\mathbf{I})$. When running KMdPCA, the degree-$2$ polynomial kernel used in Sec. \ref{sec:simu2} was adopted, and weights were set as $\omega_1=\omega_2=0.5$.

Figure \ref{fig:kmdpca} depicts the first two dPCs of KMdPCA, MdPCA, KdPCA and dPCA, as well as the first two PCs of (K)PCA.
It is evident that only KMdPCA is able to discover the two clusters in the target data.

\section{Concluding Summary}\label{sec:concl}
In diverse practical setups, one is interested in extracting, visualizing, and leveraging the unique low-dimensional features of one dataset relative to a few others. This paper put forward a novel framework, that is termed discriminative (d) PCA, for performing discriminative analytics of multiple datasets. Both linear, kernel, and multi-background models were pursued.
  In contrast with existing alternatives, dPCA is demonstrated to be optimal under certain   assumptions. 
Furthermore, dPCA is \textcolor{black}{parameter free}, and requires only one generalized eigenvalue decomposition. 
Extensive  tests using both synthetic and real data corroborated the efficacy of our proposed approaches relative to relevant prior works. 

Several directions open up for future research: i) distributed and privacy-aware (MK)dPCA implementations to cope with large amounts of high-dimensional data; ii) robustifying (MK)dPCA to outliers; 
 and iii) 
 graph-aware (MK)dPCA generalizations exploiting additional priors of the data.

\section*{Acknowledgements}
The authors would like to thank Professor Mati Wax for pointing out an error in an early draft of this paper.

\bibliographystyle{IEEEtranS}

\bibliography{pca}

% Generated by IEEEtranS.bst, version: 1.14 (2015/08/26)
\begin{thebibliography}{10}
\providecommand{\url}[1]{#1}
\csname url@samestyle\endcsname
\providecommand{\newblock}{\relax}
\providecommand{\bibinfo}[2]{#2}
\providecommand{\BIBentrySTDinterwordspacing}{\spaceskip=0pt\relax}
\providecommand{\BIBentryALTinterwordstretchfactor}{4}
\providecommand{\BIBentryALTinterwordspacing}{\spaceskip=\fontdimen2\font plus
\BIBentryALTinterwordstretchfactor\fontdimen3\font minus
  \fontdimen4\font\relax}
\providecommand{\BIBforeignlanguage}[2]{{%
\expandafter\ifx\csname l@#1\endcsname\relax
\typeout{** WARNING: IEEEtranS.bst: No hyphenation pattern has been}%
\typeout{** loaded for the language `#1'. Using the pattern for}%
\typeout{** the default language instead.}%
\else
\language=\csname l@#1\endcsname
\fi
#2}}
\providecommand{\BIBdecl}{\relax}
\BIBdecl

\bibitem{abdi2013multiple}
H.~Abdi, L.~J. Williams, and D.~Valentin, ``Multiple factor analysis: Principal
  component analysis for multitable and multiblock data sets,'' \emph{Wiley
  Interdisciplinary Reviews: Comput. Stat.}, vol.~5, no.~2, pp. 149--179, Mar.
  2013.

\bibitem{2017cpca}
A.~Abid, V.~K. Bagaria, M.~J. Zhang, and J.~Zou, ``Contrastive principal
  component analysis,'' \emph{arXiv:1709.06716}, 2017.

\bibitem{RKHS}
N.~Aronszajn, ``Theory of reproducing kernels,'' \emph{Trans. Amer. Math.
  Soc.}, vol.~68, no.~3, pp. 337--404, 1950.

\bibitem{mkl2004}
F.~R. Bach, G.~R. Lanckriet, and M.~I. Jordan, ``Multiple kernel learning,
  conic duality, and the {SMO} algorithm,'' in \emph{Proc. of Intl. Conf. on
  Mach. Learn.}, New York, USA, Jul. 4-8, 2004.

\bibitem{mhealth}
O.~Banos, C.~Villalonga, R.~Garcia, A.~Saez, M.~Damas, J.~A. Holgado-Terriza,
  S.~Lee, H.~Pomares, and I.~Rojas, ``Design, implementation and validation of
  a novel open framework for agile development of mobile health applications,''
  \emph{Biomed. Eng. Online}, vol.~14, no.~2, p.~S6, Dec. 2015.

\bibitem{barshan2011supervised}
E.~Barshan, A.~Ghodsi, Z.~Azimifar, and M.~Z. Jahromi, ``Supervised principal
  component analysis: Visualization, classification and regression on subspaces
  and submanifolds,'' \emph{Pattern Recognit.}, vol.~44, no.~7, pp. 1357--1371,
  2011.

\bibitem{2003eigenmap}
M.~Belkin and P.~Niyogi, ``Laplacian eigenmaps for dimensionality reduction and
  data representation,'' \emph{Neural Comput.}, vol.~15, no.~6, pp. 1373--1396,
  Jun. 2003.

\bibitem{asilomar2018cwg}
J.~Chen, G.~Wang, and Giannakis, ``Nonlinear discriminative dimensionality
  reduction of multiple datasets,'' in \emph{Proc. of Asilomar Conf. on
  Signals, Syst., and Comput.}, Pacific Grove, CA, USA, Oct. 28-31, 2018.

\bibitem{2018gmcca}
------, ``Graph multiview canonical correlation analysis,'' \emph{IEEE Trans.
  Signal Process.}, submitted Nov. 2018.

\bibitem{2018cwsggcca}
J.~Chen, G.~Wang, Y.~Shen, and G.~B. Giannakis, ``Canonical correlation
  analysis of datasets with a common source graph,'' \emph{IEEE Trans. Signal
  Process.}, vol.~66, no.~16, pp. 4398--4408, Aug. 2018.

\bibitem{fidler2006combining}
S.~Fidler, D.~Skocaj, and A.~Leibardus, ``Combining reconstructive and
  discriminative subspace methods for robust classification and regression by
  subsampling,'' \emph{IEEE Trans. Pattern Analysis \& Machine Intell.},
  vol.~28, no.~3, pp. 337--350, Mar. 2006.

\bibitem{1933lda}
R.~A. Fisher, ``The use of multiple measurements in taxonomic problems,''
  \emph{Ann. Eugenics}, vol.~7, no.~2, pp. 179--188, Sep. 1936.

\bibitem{2013fukunaga}
K.~Fukunaga, \emph{Introduction to Statistical Pattern Recognition}.\hskip 1em
  plus 0.5em minus 0.4em\relax 2nd ed., San Diego, CA, USA: Academic Press,
  Oct. 2013.

\bibitem{1998background}
S.~Garte, ``The role of ethnicity in cancer susceptibility gene polymorphisms:
  {T}he example of {CYP1A1},'' \emph{Carcinogenesis}, vol.~19, no.~8, pp.
  1329--1332, Aug. 1998.

\bibitem{proc2018gsk}
G.~B. Giannakis, Y.~Shen, and G.~V. Karanikolas, ``Topology identification and
  learning over graphs: {A}ccounting for nonlinearities and dynamics,''
  \emph{Proc. IEEE}, vol. 106, no.~5, pp. 787--807, May 2018.

\bibitem{guo2003generalized}
Y.~Guo, S.~Li, J.~Yang, T.~Shu, and L.~Wu, ``A generalized {F}oley--{S}ammon
  transform based on generalized {F}isher discriminant criterion and its
  application to face recognition,'' \emph{Pattern Recognit. Lett.}, vol.~24,
  no. 1-3, pp. 147--158, Jan. 2003.

\bibitem{hastie2009elements}
T.~Hastie, R.~Tibshirani, and J.~Friedman, \emph{The Elements of Statistical
  Learning 2nd edition}.\hskip 1em plus 0.5em minus 0.4em\relax New York:
  Springer, 2009.

\bibitem{mice}
C.~Higuera, K.~J. Gardiner, and K.~J. Cios, ``Self-organizing feature maps
  identify proteins critical to learning in a mouse model of down syndrome,''
  \emph{PloS {ONE}}, vol.~10, no.~6, p. e0129126, Jun. 2015.

\bibitem{1936cca}
H.~Hotelling, ``Relations between two sets of variates,'' \emph{Biometrika},
  vol.~28, no. 3/4, pp. 321--377, Dec. 1936.

\bibitem{hou2016discriminative}
C.~Hou, F.~Nie, and D.~Tao, ``Discriminative vanishing component analysis,'' in
  \emph{AAAI}, Phoenix, Arizona, USA, Fec. 12-17, 2016, pp. 1666--1672.

\bibitem{2014mati}
A.~Jaffe and M.~Wax, ``Single-site localization via maximum discrimination
  multipath fingerprinting.'' \emph{IEEE Trans. Signal Process.}, vol.~62,
  no.~7, pp. 1718--1728, April 2014.

\bibitem{1901pca}
F.~R.~S. Karl~Pearson, ``{LIII}. {O}n lines and planes of closest fit to
  systems of points in space,'' \emph{The London, Edinburgh, and Dublin Phil.
  Mag. and J. of Science}, vol.~2, no.~11, pp. 559--572, 1901.

\bibitem{cifar10}
A.~Krizhevsky, ``Learning multiple layers of features from tiny images,'' in
  \emph{Master's Thesis}, Department of Computer Science, University of
  Toronto, 2009.

\bibitem{mds}
J.~B. Kruskal, ``Multidimensional scaling by optimizing goodness of fit to a
  nonmetric hypothesis,'' \emph{Psychometrika}, vol.~29, no.~1, pp. 1--27, Mar.
  1964.

\bibitem{yaleb}
K.~C. Lee, J.~Ho, and D.~J. Kriegman, ``Acquiring linear subspaces for face
  recognition under variable lighting,'' \emph{IEEE Trans. Pattern Anal. Mach.
  Intell.}, vol.~27, no.~5, pp. 684--698, May 2005.

\bibitem{mika1999fisher}
S.~Mika, G.~Ratsch, J.~Weston, B.~Scholkopf, and K.-R. Mullers, ``Fisher
  discriminant analysis with kernels,'' in \emph{Neural Net. Signal Process.
  IX: Proc. IEEE Signal Process. Society Workshop}, Madison, WI, USA, Aug. 25,
  1999, pp. 41--48.

\bibitem{ng2002spectral}
A.~Y. Ng, M.~I. Jordan, and Y.~Weiss, ``On spectral clustering: Analysis and an
  algorithm,'' in \emph{Adv. in Neural Inf. Process. Syst.}, Vancouver, British
  Columbia, Canada, Dec. 3-8, 2001, pp. 849--856.

\bibitem{lle}
S.~T. Roweis and L.~K. Saul, ``Nonlinear dimensionality reduction by locally
  linear embedding,'' \emph{Science}, vol. 290, no. 5500, pp. 2323--2326, Dec.
  2000.

\bibitem{saad1}
Y.~Saad, \emph{Iterative {M}ethods for {S}parse {L}inear {S}ystems}.\hskip 1em
  plus 0.5em minus 0.4em\relax 2nd ed., Philadelphia, PA, USA: SIAM, 2003.

\bibitem{kpca}
B.~Scholkopf, A.~Smola, and K.~B. Muller, \emph{Kernel Principal Component
  Analysis}.\hskip 1em plus 0.5em minus 0.4em\relax Berlin, Heidelberg:
  Springer, 1997, pp. 583--588.

\bibitem{jstsp2016shahid}
N.~Shahid, N.~Perraudin, V.~Kalofolias, G.~Puy, and P.~Vandergheynst, ``Fast
  robust {PCA} on graphs,'' \emph{IEEE J. Sel. Topics Signal Process.},
  vol.~10, no.~4, pp. 740--756, Feb. 2016.

\bibitem{2004kernel}
J.~Shawe-Taylor and N.~Cristianini, \emph{Kernel Methods for Pattern
  Analysis}.\hskip 1em plus 0.5em minus 0.4em\relax Cambridge university press,
  Jun. 2004.

\bibitem{2017kpca}
J.~B. Souza~Filho and P.~S. Diniz, ``A fixed-point online kernel principal
  component extraction algorithm,'' \emph{IEEE Trans. Signal Process.},
  vol.~65, no.~23, pp. 6244--6259, Dec. 2017.

\bibitem{2000isomap}
J.~B. Tenenbaum, V.~De~Silva, and J.~C. Langford, ``A global geometric
  framework for nonlinear dimensionality reduction,'' \emph{Science}, vol. 290,
  no. 5500, pp. 2319--2323, Dec. 2000.

\bibitem{2018l1pca}
N.~Tsagkarakis, P.~P. Markopoulos, G.~Sklivanitis, and D.~A. Pados, ``L1-norm
  principal component analysis of complex data,'' \emph{IEEE Trans. Signal
  Process.}, vol.~66, no.~12, pp. 3256--3267, Mar. 2018.

\bibitem{2018dpca}
G.~Wang, J.~Chen, and G.~B. Giannakis, ``{DPCA}: {D}imensionality reduction for
  discriminative analytics of multiple large-scale datasets,'' in \emph{Proc.
  of Intl. Conf. on Acoustics, Speech, and Signal Process.}, Calgary, AB,
  Canada, April 15-20, 2018.

\bibitem{1995pca}
B.~Yang, ``Projection approximation subspace tracking,'' \emph{IEEE Trans.
  Signal Process.}, vol.~43, no.~1, pp. 95--107, Jan. 1995.

\bibitem{zhang2017going}
L.~Zhang, G.~Wang, and G.~B. Giannakis, ``Going beyond linear dependencies to
  unveil connectivity of meshed grids,'' in \emph{{IEEE} Wkshp. on Comput. Adv.
  Multi-Sensor Adapt. Process.}, Curacao, Dutch Antilles, Dec. 2017.

\bibitem{tsp2017zwrg}
L.~Zhang, G.~Wang, D.~Romero, and G.~B. Giannakis, ``Randomized block
  {F}rank--{W}olfe for convergent large-scale learning,'' \emph{IEEE Trans.
  Signal Process.}, vol.~65, no.~24, pp. 6448--6461, Dec. 2017.

\end{thebibliography}
\appendix

{\color{black}
%	\subsection{Proof of the equivalence between \eqref{eq:kdpca} and \eqref{eq:kdpcafm2o}}\label{app:kdpca}
		We start by showing that
		\begin{equation}
		\label{eq:num}
		\bm{\Phi}^\top(\mathbf{Z}) \mathbf{C}_{xx}^{\phi}\bm{\Phi}(\mathbf{Z})\mathbf{a}=\mathbf{K}\mathbf{K}^x\mathbf{a}\in\mathbb{R}^{N}.
		\end{equation}
		For notational brevity, let $\bm{\phi}_i$, $a_i$, and $K_{i,j}$ denote the $i$-th column of $\bm{\Phi}(\mathbf{Z})$, the $i$-th entry of $\mathbf{a}$, and the $(i,\,j)$-th entry of $\mathbf{K}$, respectively.
		Thus, the $i$-th element of the left-hand-side  of \eqref{eq:num} can be rewritten as
		\begin{align*}
		&\bm{\phi}_i^\top \frac{1}{m}\sum_{j=1}^m \bm{\phi}_j \bm{\phi}_j^\top \sum_{k=1}^N a_k \bm{\phi}_k
		=\frac{1}{m}\sum_{k=1}^N a_k \sum_{j=1}^m K_{i,j}K_{j,k}\\
		&=\sum_{k=1}^N a_k \sum_{j=1}^m K_{i,j} K_{j, k}^x 
		=\sum_{k=1}^N a_k \sum_{j=1}^N K_{i,j} K_{j, k}^x\\
		&=\sum_{k=1}^N a_k s_{i,k}=\mathbf{s}_i^\top\mathbf{a}=\mathbf{k}_i^\top\mathbf{K}^x\mathbf{a}
		\end{align*}
		where $s_{i,k}\in\mathbb{R}$, $\mathbf{s}_i^\top\in\mathbb{R}^{1\times N}$, and $\mathbf{k}_i^\top\in\mathbb{R}^{1\times N}$ correspond to the $(i,\,k)$-th entry of $\mathbf{K}\mathbf{K}^x$, the $i$-th row of $\mathbf{K}\mathbf{K}^x$, and the $i$-th row of $\mathbf{K}$, respectively. Evidently, $\mathbf{k}_i^\top\mathbf{K}^x\mathbf{a}$ is the $i$-th entry of the right-hand-side in \eqref{eq:num}, which proves that \eqref{eq:num} holds. It follows that the numerators of \eqref{eq:kdpca} and \eqref{eq:kdpcafm2o} are identical.
		
		Similarly, one can verify that
		\begin{equation}
		\label{eq:den}
		\mathbf{a}^\top\bm{\Phi}^\top(\mathbf{Z}) \mathbf{C}_{yy}^{\phi}\bm{\Phi}(\mathbf{Z})\mathbf{a}=\mathbf{a}^\top\mathbf{K}\mathbf{K}^y\mathbf{a}.
		\end{equation}
		Hence, \eqref{eq:kdpca} and \eqref{eq:kdpcafm2o} are equivalent.

}

\end{document}